\documentclass{article}

\usepackage{microtype}
\usepackage{graphicx}
\usepackage{subcaption}
\usepackage{amsmath}
\usepackage{amssymb}
\usepackage{algorithm}
\usepackage{algorithmic}
\usepackage{booktabs} 

\usepackage{hyperref}

\usepackage[preprint]{icml2026}

\usepackage{amsmath}
\usepackage{amssymb}
\usepackage{mathtools}
\usepackage{amsthm}

\usepackage{dirtree}
\usepackage{xcolor}
\definecolor{pyblue}{RGB}{30,100,200}
\definecolor{mdred}{RGB}{180,60,0}
\definecolor{jsongreen}{RGB}{0,130,60}

\usepackage[capitalize,noabbrev]{cleveref}

\theoremstyle{plain}

\theoremstyle{definition}

\theoremstyle{remark}

\usepackage[textsize=tiny]{todonotes}

\icmltitlerunning{Lang2MLIP}

\begin{document}

\twocolumn[
  \icmltitle{
Lang2MLIP: End-to-End Language-to-Machine Learning Interatomic Potential Development with Autonomous Agentic Workflows}



  \icmlsetsymbol{equal}{*}

  \begin{icmlauthorlist}
    \icmlauthor{Wenwen Li}{comp}
    \icmlauthor{Yuki Orimo}{comp}
    \icmlauthor{Nontawat Charoenphakdee}{comp}
  \end{icmlauthorlist}

  \icmlaffiliation{comp}{Preferred Networks, Inc., Tokyo, Japan}

  \icmlcorrespondingauthor{Wenwen Li}{wenwenli@preferred.jp}

\vskip 0.3in
]



\printAffiliationsAndNotice{}  

\begin{abstract}

Developing machine learning interatomic potentials (MLIPs) for complex materials systems remains challenging because it requires expertise in atomistic simulations, machine learning, and workflow design, as well as iterative active learning procedures. 
Existing automated pipelines typically assume a fixed sequence of stages or depend on domain experts, which limits their adaptability to heterogeneous materials systems where the optimal curriculum is not known in advance.
To lower the barrier to developing MLIPs for non-experts, we propose \textbf{Lang2MLIP}, a multi-agent framework that takes natural-language input and formulates end-to-end MLIP development as a sequential decision-making problem solved by large language models (LLMs).
At each step, a decision-making agent observes the current dataset, model, evaluation results, and execution log, and then automatically selects an appropriate action to improve the model. 
This removes the need for a predefined pipeline and enables the agent to self-correct by revisiting earlier subsystems when new failures arise.
We evaluate this approach on a solid electrolyte interphase (SEI) system with multiple components and interfaces. 
These results suggest that LLM-based multi-agent systems are a promising direction for automating MLIP development and making it more accessible to non-experts.
\end{abstract}
\section{Introduction}
Machine learning interatomic potentials (MLIPs) have become an important tool for atomistic simulation, offering near first-principles accuracy at substantially lower computational cost than \emph{ab initio} methods \citep{bpnet,GAP,ace,behler2016perspective,deringer2019machine,unke2021machine}. 
Recent advances in model architectures have further broadened the scope of MLIPs across molecules, materials, and interfaces \citep{schnet,mace,nequip,m3gnet,chgnet,takamoto2022towards,zhang2024dpa,mattersim,rhodes2025orb}. 

Despite recent progress, developing an MLIP for a complex materials system still requires domain expertise across many stages, including structure design, simulation protocol engineering, error analysis, and active learning, as well as repeated debugging and workflow refinement~\citep{zuo2020performance,morrow2022indirect,dai2025uncertainty}. 
These challenges are even more pronounced in heterogeneous materials systems, where diverse chemical species and local environments complicate data generation and model validation. 
A representative example is the battery solid electrolyte interphase (SEI)~\cite{wang2018review,li2023review,wu2025formation}, which contains chemically distinct components and interfacial regions whose local environments can change substantially during simulation. 
Training data for such systems must cover bulk-like structures, disordered configurations, interfaces, and thermally perturbed states. 
Existing MLIP workflows for these settings typically rely on manually designed pipelines and fixed active learning heuristics, making them difficult for non-experts to use \citep{mtp,gong2025predictive,dpa2-workflow-wang2025pre,lahouari2025automated}.

In this paper, we present \textbf{Lang2MLIP}, a language-driven multi-agent framework that formulates MLIP development as a sequential decision problem solved by a tool-using LLM agent. 
Given only a natural-language task specification, the agent must decide at each step which action to take, how to parameterize it, and when to terminate. 
We identify workflow orchestration as the primary bottleneck in practical MLIP development.
Lang2MLIP divides the overall process into two phases. 
In the \emph{interactive preparation phase}, specialized agents interpret the task description, request missing information from the user when needed, generate initial structures, and optionally prepare a reference molecular dynamics (MD) workflow. 
In the \emph{autonomous training phase}, a central decision-making agent coordinates iterative model development at each iteration by selecting from a set of actions: sampling, training, dataset selection, evaluation, pruning, reference calculation, and termination.
Instead of following a fixed active-learning schedule, the agent chooses each action based on the current dataset, model state, evaluation results, and execution logs.
Unless termination is selected, the decision-making agent then initiates interaction with the corresponding action-specific agent to execute the chosen operation.

We evaluate Lang2MLIP on a solid electrolyte 
interphase (SEI) system in lithium batteries with
four layers, six components, and six elements, a setting whose multi-scale structure
makes the optimal curriculum difficult to specify in advance. From natural-language interactions, the agent autonomously constructs a three-stage
curriculum that progresses from basic components through binary interfaces to the
full multilayer SEI, reallocating sampling effort in response to observed model
failures. These results provide evidence that LLM-based sequential decision making
is a viable substitute for hand-engineered active-learning pipelines in a non-trivial
scientific setting.


\section{Related Work}
\subsection{Training Dataset Preparation in MLIP Development}
A central challenge in MLIP development is constructing training datasets that
adequately cover the relevant configuration space, since insufficient coverage can
lead to poor transferability or unreliable predictions~\citep{mtp,zuo2020performance,
morrow2022indirect,dai2025uncertainty}. 
In practice, dataset preparation is therefore
iterative, involving structure generation, model evaluation, and retraining. Active
learning is a common strategy for improving sample efficiency, typically by alternating
between candidate generation, informative sample selection based on uncertainty or error
signals, and retraining~\citep{mtp,gong2025predictive,dpa2-workflow-wang2025pre,
vinod2025lfab}. 
These pipelines are typically structured as a fixed sequence of
stages with hand-tuned criteria, and the selection among them is usually made by a
human expert.

\subsection{LLM for Scientific Discovery and Simulation}

Recent work has explored LLMs as autonomous agents for scientific discovery, where they interact with tools, reason over intermediate outputs, and execute multi-step workflows from natural language instructions. Early systems such as ChemCrow~\citep{m2024augmenting} and Coscientist~\citep{boiko2023autonomous} showed that tool-augmented LLMs can support or automate chemistry tasks ranging from retrosynthesis to robotic experimentation. In molecular simulation, related frameworks have been developed for molecular dynamics, polymer simulation, density functional theory (DFT), and quantum chemistry \citep{campbell2026mdcrow,zhao2026polyjarvis,orimo2025parc,wang2025dreams,liu2025vaspilot,zou2025agente,shi2025fine,wei2025ai}.

These systems demonstrate the potential of LLM-based agents for scientific workflows, but mainly focus on simulation setup and execution. 
In contrast, our work targets the iterative workflow of MLIP development, which requires dataset construction, adaptive sampling, and repeated model refinement.

\subsection{Automatic Frameworks for MLIP development}

Closest to our work are recent frameworks that connect natural-language instructions to trained MLIPs. 
AMLP~\citep{lahouari2025automated} uses a multi-agent LLM system to automate DFT setup, \emph{ab initio} molecular dynamics sampling, and fine-tuning of a MACE foundation model from a short user prompt. 
Nevertheless, its automation is primarily focused on task execution within a human-driven active learning framework, where users still specify key decisions such as simulation conditions, filtering thresholds, initial structure, and training hyperparameters. 
QUASAR~\citep{yang2026quasar} uses a three-agent architecture that coordinates DFT, MLIP training, and molecular dynamics in a unified pipeline. 
PFD~\citep{dpa2-workflow-wang2025pre} introduces a fixed pre-training, fine-tuning, and distillation workflow that generates material-specific MLIPs from a pretrained universal model, but without an LLM-based controller.

\begin{figure*}
  \centering
  \includegraphics[width=1.8\columnwidth]{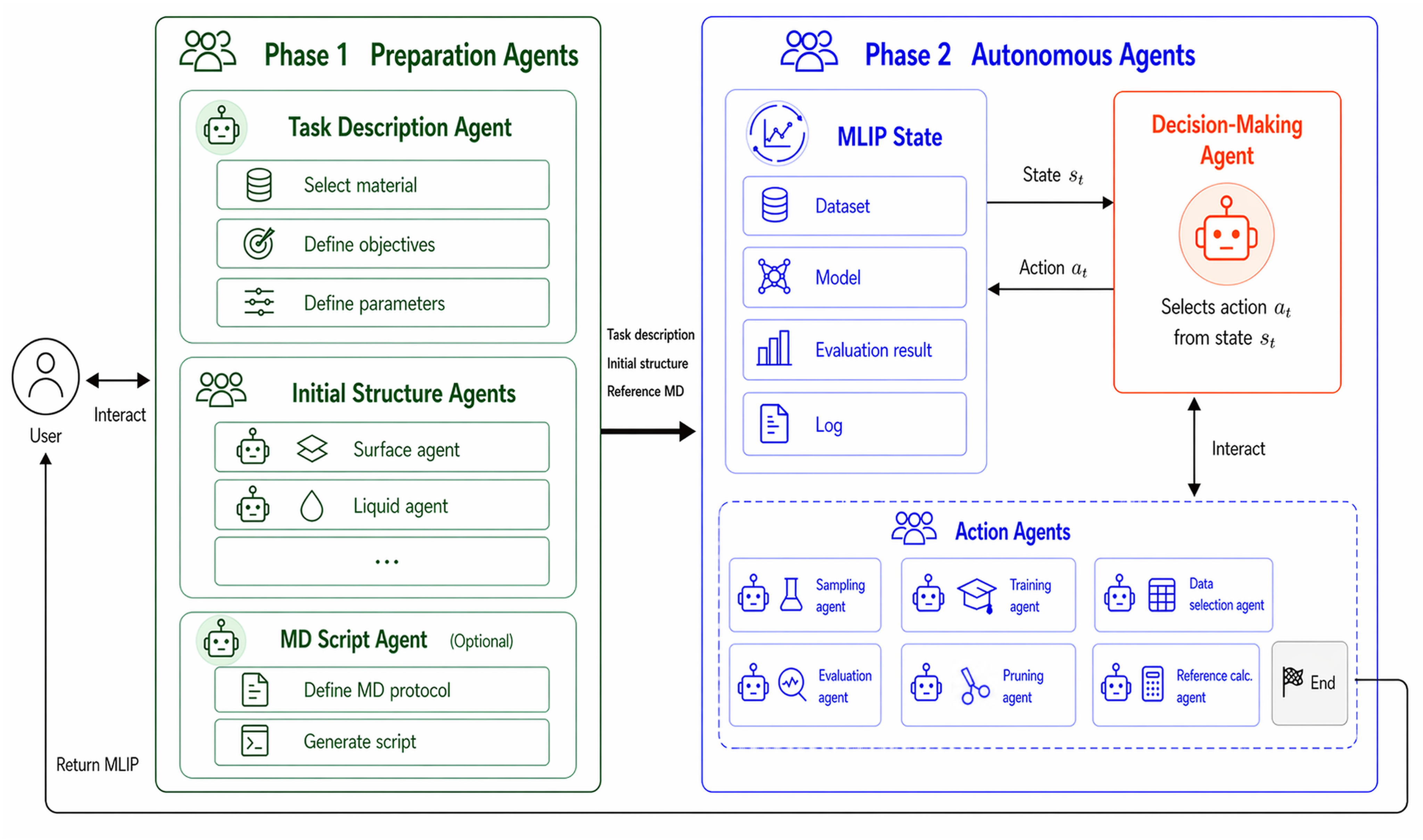} %
  \caption{Overview of the Lang2MLIP framework. The robot icon indicates a single agent, while the group-of-people icon indicates multiple agents within a submodule.}
  \label{fig:architecture}
\end{figure*}

Lang2MLIP differs from prior automated MLIP pipelines in a structural way.
Existing systems such as AMLP~\citep{lahouari2025automated} and
QUASAR~\citep{yang2026quasar} couple LLM agents to a predefined sequence of stages
(parameter recommendation, input generation, QM execution, dataset formatting,
training) and retain a human in the loop for active-learning decisions. 
Lang2MLIP removes this dependency. Once the task is clarified, the autonomous phase selects actions via an LLM conditioned on the current state, operating without further human involvement.
Our central premise is that workflow-level coordination over a set of actions is sufficient to support effective autonomous MLIP development, and that
the resulting behavior, including self-correction and adaptive curriculum construction,
emerges from decision making rather than from hand-engineered stage transitions.

\section{Lang2MLIP Framework}
\subsection{Overview}
Figure~\ref{fig:architecture} provides an overview of Lang2MLIP, a multi-agent framework with two operational phases: an \emph{interactive preparation phase} and an \emph{autonomous training phase}. 

In the first phase, specialized agents gather task information, generate initial structures, and, optionally, prepare reference MD scripts for advanced simulation settings (e.g., non-equilibrium MD or metadynamics). 
This stage supports human-in-the-loop interaction, allowing users to inspect and refine task specifications with language.

In the second phase, agents collaboratively and iteratively improve the MLIP under the control of a central decision-making agent. 
At each step, this agent assesses the current state of the model and selects the most appropriate action to improve performance. 
The process continues until the model meets the desired criteria, at which point the workflow terminates and returns the MLIP to the user. 
This phase is designed for fully automated, long-running execution, encompassing training, failure analysis, data prioritization, and iterative refinement without manual intervention.

\subsection{Multi-Agent Architecture}
All agents are defined via role-specific prompts using the Claude Agent SDK~\footnote{\url{https://code.claude.com/docs/en/agent-sdk/overview}} with Claude Opus 4.6~\footnote{\url{https://www.anthropic.com/news/claude-opus-4-6}}.
For example, the decision-making agent is prompted as a top-level controller responsible for determining and outputting the next task to execute (see Appendix~\ref{app:decision-maker} for a full prompt).

\subsubsection{Interactive Preparation Phase}
This phase consists of three specialized agents shown in Figure~\ref{fig:architecture} (left): Task Description agent, Initial Structure Agents, MD Script Agent, where  Initial Structure Agents have several sub-agents.
This phase is interactive because user intent and scientific scope are often ambiguous at the start of an MLIP project, and iterative clarification is essential for defining a well-posed training objective.

\textbf{Task Description Agent} identifies the material system, simulation objectives, and task parameters through interactive dialogue with the user.

\begin{table}
\centering
\caption{Initial structure generation agents}
\label{tab:structure_gen_agents}
\begin{tabular}{@{}p{0.32 \columnwidth} p{0.66\columnwidth}@{}}
\toprule
\textbf{Agent} & \textbf{Scope} \\
\midrule
\texttt{Solid} & Crystalline bulk solids, including solid solutions. \\
\texttt{Amorphous} & Non-crystalline inorganic materials (e.g., glasses and melts). \\
\texttt{Molecule} & Organic molecular systems such as liquids, mixtures, and small molecules. \\
\texttt{Solid-surface} & Crystal surfaces and slab models. \\
\texttt{Cluster} & Finite clusters or nanoparticles (non-periodic). \\
\texttt{Solid-solid} & Interfaces between solid phases. \\
\texttt{Solid-molecule} & Solid-molecule interfaces and adsorption. \\
\texttt{Liquid-liquid} & Interfaces between molecular liquids. \\
\texttt{Polymer} & Long-chain molecules (e.g., polymers). \\
\texttt{Other} & Other structures. \\
\bottomrule
\end{tabular}
\end{table}

\textbf{Initial Structure Generation Agents} are crucial because initial dataset quality and coverage strongly influence downstream MLIP performance. 
The agent first analyzes the task to identify the most informative configuration types for training. 
For systems with solid--liquid interfaces, for instance, it considers bulk solid, bulk liquid, and interfacial configurations. 
This decomposition matters because MLIPs learn from local atomic environments, e.g., interfacial regions exhibit coordination patterns not well represented by bulk phases alone. Based on this analysis, the agent coordinates a set of sub-agents (see Table~\ref{tab:structure_gen_agents}) to construct candidate structures, systematically varying parameters such as lattice size, density, composition, and interfacial geometry, and applying reusable utility functions to improve diversity and physical plausibility. 
The process is interactive: the agent queries the user for missing or task-specific information (e.g., Materials Project IDs, SMILES strings, composition details, or simulation constraints), enabling context-aware generation without requiring fully specified inputs upfront.

\textbf{MD Script Generation Agent} is optionally used for complex simulation scenarios . 
While standard ensembles (e.g., NVT or NPT) can be managed reliably by downstream agents, more advanced protocols such as non-equilibrium molecular dynamics or meta-dynamics, remain difficult to specify correctly within the autonomous training phase. 
To address this limitation, the agent produces specialized reference MD scripts during the preparation stage, allowing users to validate them before execution. 
Once verified, these scripts guide subsequent sampling during MLIP training and improve coverage of configurations relevant to the target simulation conditions.

The outputs of the interactive preparation phase are (i) a structured task description, (ii) a set of initial structures, and (iii) an optional reference MD script. 

\subsubsection{Autonomous training phase}
Once the preparation phase is completed, the workflow enters an autonomous phase consisting of a decision-making agent and several action agents responsible for sampling, training, evaluation, and data management.

\begin{table}
\centering
\caption{Available actions for the decision-making agent.}
\label{tab:action_space}
\begin{tabular}{@{}p{0.33\columnwidth} p{0.62\columnwidth}@{}}
\toprule
\textbf{Action} & \textbf{Description} \\
\midrule
\texttt{Sample}         & Generate new configurations using a reference (e.g.\ DFT) or the current MLIP model. \\
\texttt{Select\_data}   & Select informative structures from the pool and label them \\
\texttt{Train}          & Train or fine-tune MLIP on selected datasets. \\
\texttt{Model\_eval}    & Evaluate MLIP on validation tasks. \\
\texttt{Prune}          & Remove redundant components or low-quality configurations. \\
\texttt{Reference\_calc}& Run the reference method on validation tasks for future MLIP evaluation. \\
\texttt{End}            & Finish: return MLIP and final report. \\
\bottomrule
\end{tabular}
\vspace{-7mm}
\end{table}

\textbf{The Decision-making Agent} is the central controller of the autonomous workflow. 

The process can be formalized as a sequential decision problem.
At each step $t$, the decision-making agent observes a state
$s_t = (\mathcal{D}_t, \mathcal{M}_t, \mathcal{E}_t, \mathcal{L}_t)$,
where $\mathcal{D}_t$ denotes the current dataset, $\mathcal{M}_t$ the current MLIP,
$\mathcal{E}_t$ the accumulated evaluation results (e.g.\ validation errors,
stability diagnostics, comparison of basic properties with reference model), and $\mathcal{L}_t$ a log of
prior commands, agent-agent discussions, and any failure signals.
It then decides the next action, where possible actions are outlined in Table~\ref{tab:action_space}.
After executing the action $a_t$,
the system updates its artifacts (e.g., sampled trajectories, selected structures,
models, or evaluation results) and moves to the next state
$s_{t+1}$. 
The workflow terminates when the decision-making agent selects \texttt{End}.

Once an action is selected, the decision-making agent interacts with the corresponding \textbf{action agents} (outlined in Table~\ref{tab:action_space}) to execute it. 
For example, if sampling is selected, the decision-making agent provides the sampling agent with the task description and relevant parameters needed to carry out the job.




\section{Experimental Results}

In this section, we demonstrate the effectiveness of Lang2MLIP for a battery SEI. 
In this experiment, we adopt the Preferred Potential (PFP), a universal interatomic potential, as the reference data source due to its high reliability and significantly lower computational cost compared to DFT \cite{takamoto2022towards}. 
The target MLIP is based on the moment tensor potential (MTP) architecture~\citep{mtp}. 
We note that both the choice of reference data source and the MLIP architecture are flexible and can be readily replaced by other alternatives. 
The only requirement is that the corresponding simulation workflows such as single-point energy calculations, MD simulations and model training command, are properly exposed and executable by the agents.

\begin{table*}
\centering
\caption{Action log for the Lang2MLIP workflow on the multilayer SEI system. See Appendix ~\ref{app:workflow_log} for details.}
\label{tab:workflow_log}
\small
\setlength{\tabcolsep}{5pt}
\renewcommand{\arraystretch}{1.1}
\begin{tabular}{c l p{0.73\textwidth}}
\toprule
\textbf{Step} & \textbf{Action} & \textbf{Description} \\
\midrule
1  & \texttt{Reference\_calc} & Use PFP to run MD on validation structures for reference. \\
2  & \texttt{Sample}          & Use PFP to sample all non-validation structures. \\
3  & \texttt{Train}           & Train the first MLIP model using 3312 snapshots. \\
4  & \texttt{Sample}          & Use the current MLIP to sample basic components (graphite, Li$_2$CO$_3$, LEDC, electrolyte). \\
5  & \texttt{Select\_data}    & Select the top 7.5\% high-error snapshots. \\
6  & \texttt{Train}           & Fine-tune the model with 364 high-error samples. \\
7  & \texttt{Sample}          & Use the current MLIP to sample interface structures (graphite/Li$_2$CO$_3$, Li$_2$CO$_3$/LEDC). \\
8  & \texttt{Select\_data}    & Select high-error samples, prioritizing graphite/Li$_2$CO$_3$. \\
9  & \texttt{Train}           & Fine-tune the model with 1457 interface samples. \\
10 & \texttt{Sample}          & Use the current MLIP to re-sample interfaces at improved high temperature. \\
11 & \texttt{Select\_data}    & Select 7.5\% high-error samples. \\
12 & \texttt{Train}           & Fine-tune the model with 1400 high-error interface samples. \\
13 & \texttt{Model\_eval}     & Evaluate the model on interface structures by comparison with PFP. \\
14 & \texttt{Sample}          & Use the current MLIP to sample LEDC/electrolyte interface structures. \\
15 & \texttt{Select\_data}    & Select 7.5\% high-error samples. \\
16 & \texttt{Train}           & Fine-tune the model with 812 LEDC/electrolyte interface samples. \\
17 & \texttt{Sample}          & Use the current MLIP to sample the full multilayer SEI (graphite/Li$_2$CO$_3$/LEDC/electrolyte). \\
18 & \texttt{Select\_data}    & Select high-error samples from four-layer SEI interface snapshots. \\
19 & \texttt{Train}           & Fine-tune the model with 181 four-layer SEI samples. \\
20 & \texttt{Model\_eval}     & Evaluate the multilayer SEI model. \\
21 & \texttt{End}             & Workflow completed: return MLIP. \\
\bottomrule
\end{tabular}
\end{table*}

\subsection{Battery Solid Electrolyte Interphase (SEI)}

\begin{figure}
  \includegraphics[width=\columnwidth]{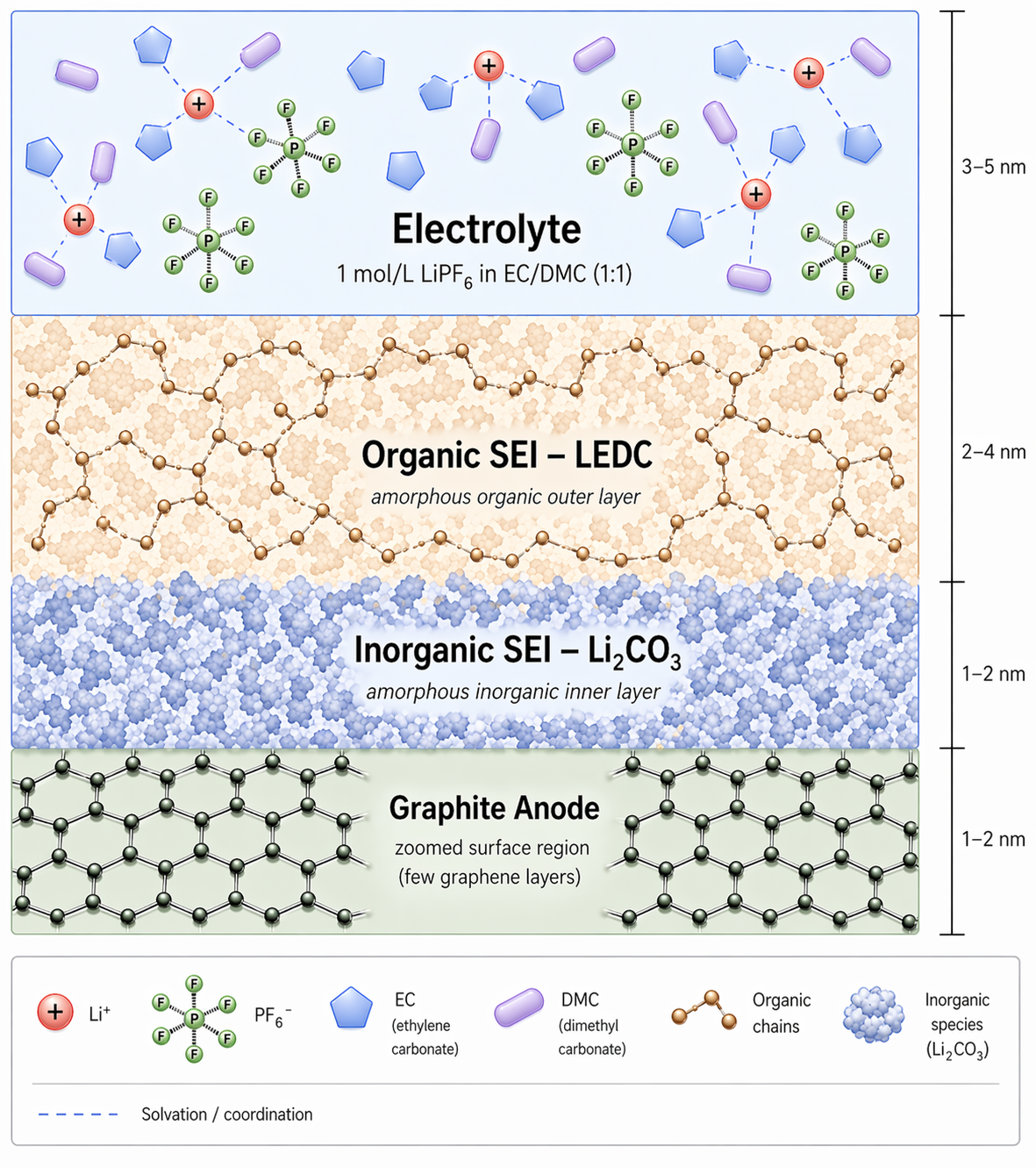}
  \caption{Schematic of the battery SEI system used in this study. 
  }
  \label{fig:sei}
\end{figure}

As illustrated in Fig. \ref{fig:sei}, the system exhibits a well-defined \textbf{four-layer hierarchical structure}. Starting from the top, the electrolyte region consists of a liquid solution (1 mol/L LiPF$_6$ in EC/DMC). 
Beneath the electrolyte lies the organic SEI outer layer, primarily composed of amorphous lithium ethylene dicarbonate (LEDC). 
This layer is relatively porous and mechanically soft, allowing partial transport of ions and solvent molecules. Below this is the inorganic SEI inner layer, dominated by amorphous inorganic compounds Li$_2$CO$_3$, which is denser and plays a critical role in electronically insulating the electrode while enabling Li$^+$ transport. 
Finally, the SEI is supported by the graphite anode, consisting of several stacked graphene layers forming the basal planes. This crystalline region provides the host structure for lithium intercalation.
Overall, the system involves organic and inorganic components, amorphous and crystalline phases, and multiple heterogeneous interfaces. Such structural and chemical heterogeneity makes it significantly more challenging to model than conventional single-phase bulk materials, particularly for the construction of a unified MLIP.

\subsection{Input Prompts}
The initial user prompt is intentionally brief, i.e. “SEI of lithium battery.” 
This prompt initiates the interactive preparation phase, during which the agent engages the user through targeted clarification questions to refine the system specification. 
These questions cover the system components, relevant phases, simulation methods and conditions, and the properties of interest. 
Once the agent determines that sufficient information has been collected, the workflow automatically transitions to the autonomous training stage, where the MLIP is developed without further human intervention. The initial prompt and the subsequent user–agent interaction process are provided in the Appendix~\ref{app:input_prompt}.


\subsection{Initial Structure Generation}

Input structure generation constitutes the most critical and challenging component of the preparation stage. To address the complexity of the SEI system, multiple agents collaboratively generate a diverse set of candidate structures spanning different levels of structural complexity. Specifically, the generated structures cover (i) individual components, including graphite bulk, graphite $(10\bar{1}0)$ surface surfaces, amorphous Li$_2$CO$_3$, amorphous LEDC, EC/DMC electrolyte, and LiPF$_6$-EC-DMC systems; (ii) pairwise interfaces, such as Li$_2$CO$_3$/LEDC, graphite $(10\bar{1}0)$ surface/Li$_2$CO$_3$, and LEDC/electrolyte interfaces; and (iii) the complete multilayer SEI assembly.

In total, ten distinct structural categories are constructed, each corresponding to a specific phase or interfacial configuration. For each category, the agents further explore structural variations by systematically modifying parameters such as system size, density, and composition, resulting in 5–7 variants per category. Altogether, 60 initial structures are generated;
details are provided in Appendix~\ref{app:init_structures}.  This hierarchical and diversified generation process ensures broad coverage of the relevant configurational space, which is essential for robust MLIP training. 
In addition, one structure from each category is selected as a validation structure for subsequent MLIP evaluation. To further understand the effect of reasonable initial structures for robust MLIP training, ablation studies are conducted in Section~\ref{sec:ablation}.

\subsection{Autonomous Training Workflow}

After the preparation agents summarize the task and generate the initial structures, the workflow enters the second stage, which is coordinated by a decision-making agent overseeing seven specialized sub-agents.

In this experiment, a total of 21 steps were carried out, with a concise summary provided in Table 3. Throughout the process, the agent performed six iterations of MLIP training, interleaved with operations such as data sampling and model evaluation. At the end of the workflow, the system produces a ready-to-use MTP model, along with corresponding validation results. 
Overall, this procedure closely resembles an active learning paradigm, while offering greater flexibility in decision-making and workflow adaptation.

\subsection{Ablation Studies}
\label{sec:ablation}

To better understand which components of Lang2MLIP are responsible for the final performance, we conduct two ablation experiments on the SEI system. 
The first ablation eliminates the use of Lang2MLIP, using manually designed initial structures and a fully manual training process.
The second ablation retains the agent-generated initial structures, but removes the autonomous closed-loop training procedure and replaces it with a simplified one-shot sampling-and-training workflow. 
These two settings allow us to separately examine the roles of the preparation phase and the autonomous training phase.

\paragraph{Ablation A: Manual initial structures without agent-assisted preparation.}
In the first ablation, we replace the preparation phase with a manually constructed initialization. 
Specifically, we create five initial structures corresponding to basic components of the SEI system, namely graphite bulk, graphite $(10\bar{1}0)$ surface surfaces, amorphous Li$_2$CO$_3$, amorphous LEDC and electrolyte. 
This choice captures the primary phases of the SEI system and is therefore physically meaningful.
Such an approach reflects a setting with limited prior experience in MLIP construction: 
assembling a small set of known component structures and relying on MD sampling to generate additional configurations, without explicitly considering interfacial diversity or coverage.
Starting from these structures, we perform sampling and subsequent MLIP training using the same reference potential as in the main experiment, but without the agent-driven preparation workflow and without the autonomous iterative refinement process. 
The total number of sampled structures is 3200, and the resulting model is evaluated on the same validation protocol as the full Lang2MLIP workflow.

\textbf{Result} During MD simulation, the trained MLIP exhibits severe instability: the system density rapidly decreases to nearly zero, and the generated structures collapse after a few picoseconds. 
This setting leads to a clear failure case. 

These results show that a small manually designed set of basic structures is inadequate for the chemically and structurally heterogeneous SEI system, leading to poor configurational coverage and unstable downstream simulations. More broadly, the main challenge is not only training the potential, but constructing a sufficiently diverse and physically plausible starting configuration. In this context, the agent-assisted preparation phase helps lower the barrier to obtaining a trainable and stable MLIP setup, especially for users who may not know in advance which structures, interfaces, and compositions to include.


\paragraph{Ablation B: Agent-generated initialization without autonomous iterative refinement.}
In the second ablation, we keep the agent-generated initial structures from the preparation phase but remove the autonomous training loop. 
Concretely, the workflow starts from the same initial structures as the full Lang2MLIP pipeline, then applies only a simplified sampling-and-training procedure without iterative model evaluation, error-driven data selection, or adaptive resampling.
The resulting training set contains 3312 structures, and the trained model is evaluated by comparing density, mean squared displacement (MSD), and radial distribution function (RDF) against the PFP reference and the Lang2MLIP model.

\begin{figure}[t]
  \includegraphics[width=\columnwidth]{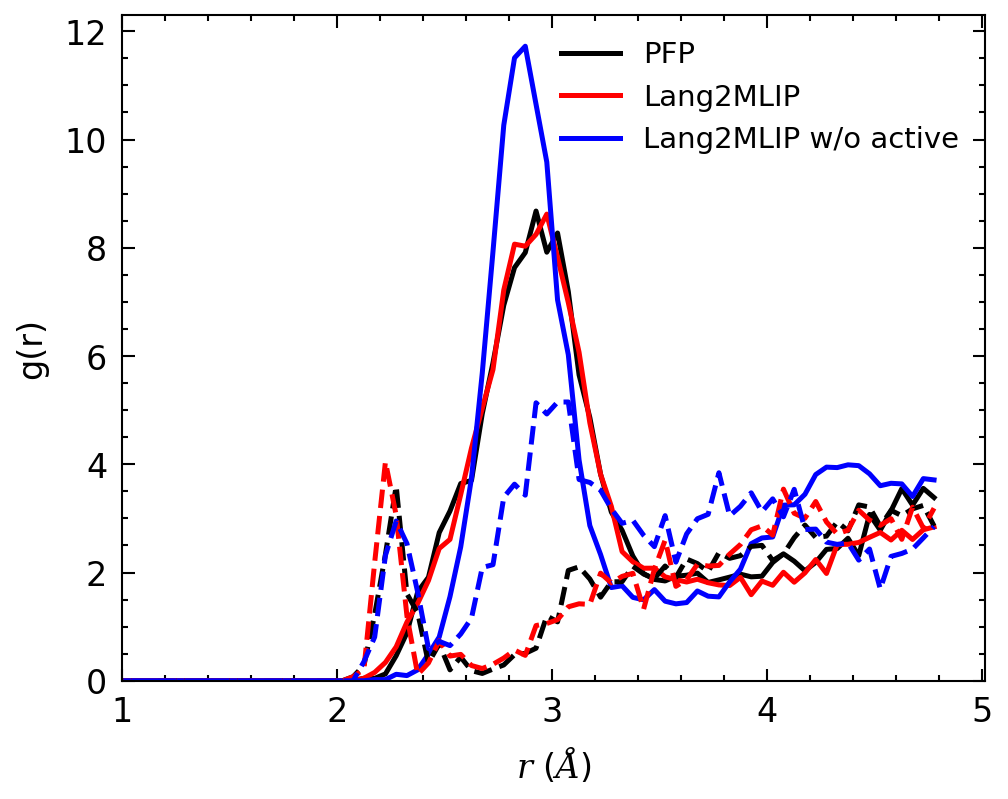}
  \caption{
  Radial distribution functions (RDFs) for selected atom pairs in the multilayer SEI system at 300 K. Solid lines show the Li–C pair in Li\textsubscript{2}CO\textsubscript{3}; dashed lines show the O–F pair in the electrolyte.
  }
  \vspace{-7mm}
  \label{fig:ablation_density_msd_rdf}
\end{figure}

\textbf{Result} Compared with Ablation A, this setting yields a substantially more stable model and physical property can be calculated. 
Figure~\ref{fig:ablation_density_msd_rdf} and Table~\ref{tab:ablation_summary} show the comparison with full Lang2MLIP method, where PFP is a reference method.
MD simulations run normally without catastrophic structural collapse, and the predicted density agrees reasonably well with the PFP reference, reaching 1.500 g/cm3 compared with 1.485 g/cm3. 
However, clear discrepancies remain in more sensitive structural and dynamical observables. 
In particular, the mean squared displacement (MSD) of Li, C, and O atoms indicates one order of magnitude slower diffusion than in PFP.
Moreover, the radial distribution functions (RDFs) still exhibit visible errors, especially for inorganic SEI and electrolyte parts. 
In contrast, the full Lang2MLIP model achieves closer agreement with PFP on these observables, as shown in Fig.~\ref{fig:ablation_density_msd_rdf} and Table~\ref{tab:ablation_summary}.
These results suggest that high-quality initialization alone is not sufficient for accurate MLIP construction, and that iterative refinement is necessary to improve quantitative fidelity in local structure and atomic dynamics. 
In this sense, the autonomous training agents fully automate iterative refinement, eliminating the need for expert-designed workflows such as active learning setup, and thereby lowering the barrier to reaching expert-comparable MLIP quality.
Full validation results are given in Appendix ~\ref{app:val_ablation}. 

\textbf{Discussion}
Taken together, the two ablations reveal a complementary division of labor between the two stages of Lang2MLIP. 
Without the preparation phase, the resulting MLIP cannot maintain stable MD trajectories, showing that initialization quality is a prerequisite for successful training. 
With agent-generated initialization but without autonomous closed-loop refinement, the model becomes stable but still underperforms on dynamical and local structural observables. 
Only the full Lang2MLIP workflow achieves both robust simulation stability and closer agreement with PFP across density, MSD, and RDF metrics.

These findings suggest that the benefit of Lang2MLIP does not arise from a single component in isolation. 
Instead, the preparation phase primarily determines whether the training problem is well-posed, whereas the autonomous training phase determines whether the final potential becomes quantitatively faithful. 
This complementary effect is especially important in complex interfacial systems such as SEI, where both broad configurational coverage and targeted iterative refinement are required.







\begin{table}[t]
\centering
\caption{Ablation results on the SEI system. Diffusion coefficients are computed from the MSD curves. PFP is the reference method. 
The Ablation A method is omitted due to instability during MD simulations, which prevents reliable evaluation of physical properties. Lang2MLIP w/o active corresponds to method in Ablation B.}
\label{tab:ablation_summary}
\small
\setlength{\tabcolsep}{4pt}
\renewcommand{\arraystretch}{1.15}
\begin{tabular}{lccc}
\toprule
\textbf{Metric} 
& \shortstack{\textbf{Lang2MLIP} \\ \textbf{w/o active}} 
& \textbf{Lang2MLIP} 
& \textbf{PFP (ref.)} \\
\midrule

Density (g/cm$^3$) 
& \textit{$1.500$} 
& \textit{$1.477$} 
& \textit{$1.485$} \\

$D_{Li}$ (cm$^2$/s) 
& \textit{$-4.61 \times 10^{-7}$} 
& \textit{$8.00 \times 10^{-7}$} 
& \textit{$7.49 \times 10^{-7}$} \\

$D_{O}$ (cm$^2$/s) 
& \textit{$4.16 \times 10^{-7}$} 
& \textit{$2.58 \times 10^{-6}$} 
& \textit{$2.29 \times 10^{-6}$} \\

$D_{C}$ (cm$^2$/s) 
& \textit{$1.3 \times 10^{-7}$} 
& \textit{$1.3 \times 10^{-6}$} 
& \textit{$1.11 \times 10^{-6}$} \\

$D_{P}$ (cm$^2$/s) 
& \textit{$8.55 \times 10^{-6}$} 
& \textit{$1.03 \times 10^{-5}$} 
& \textit{$1.59 \times 10^{-5}$} \\

\bottomrule
\end{tabular}
\end{table}

\subsection{Emergent Adaptive Behaviors}

Beyond overall task completion, we analyze the execution trace in Table~\ref{tab:workflow_log} to understand how the agent organizes the MLIP development process. Despite operating over an action space without predefined stage transitions, the system exhibits several non-trivial adaptive behaviors that resemble expert-designed workflows.

\textbf{Emergent curriculum over increasing complexity}
The agent autonomously organizes training into a three-stage progression: (i) basic components (graphite, Li$_2$CO$_3$, LEDC, electrolyte), (ii) pairwise interfaces (e.g., graphite/Li$_2$CO$_3$, Li$_2$CO$_3$/LEDC), and (iii) the full multilayer SEI system. This progression is not explicitly predefined, but emerges naturally from the sequential decision process, moving from simpler subsystems to more complex interfacial and multicomponent configurations -- a strategy that reflects expert-level intuition in MLIP development~\citep{li2025lightpfp,liang2025nep89}.

\textbf{Failure-driven data acquisition}
The agent allocates sampling effort based not only on error magnitude but also on the type and location of failure. During interface sampling (step 7), Li$_2$CO$_3$/LEDC systems remain stable across 300--800 K, whereas graphite/Li$_2$CO$_3$ simulations exhibit frequent instabilities, with 14 out of 30 runs terminating early due to structural collapse.
In response, the agent prioritizes graphite/Li$_2$CO$_3$ by allocating a larger fraction of selected high-error samples (17.5\% vs. 7.5\%) at step 8. This behavior indicates that the agent distinguishes between subsystems and concentrates resources on those that limit model robustness, rather than treating all regions uniformly.

\textbf{Adaptive exploration and iterative refinement}
The workflow exhibits iterative cycles of sampling, selection, and retraining, with the agent revisiting previously explored subsystems as new errors emerge. For example, interface configurations are refined across multiple iterations, and sampling conditions are systematically adjusted (e.g., higher-temperature sampling at step 10 than step 7) to improve coverage of challenging regions. In addition, the agent transitions from reference-based sampling in early stages to MLIP-driven sampling in later stages, reflecting increasing reliance on the learned model. These patterns indicate that the agent adapts not only where to sample, but also how sampling is performed through evolving simulation conditions and protocols.

\textbf{Efficiency of autonomous workflow construction}
The resulting workflow achieves a stable MLIP in six training iterations using 7,526 structures.
The agent determines when to advance based on model reliability, progressing to more complex subsystems once sufficient accuracy is reached on simpler ones. 
This avoids unnecessary over-optimization of intermediate components and reduces redundant computation.
As a result, the workflow allocates effort dynamically and advances in a more time-efficient manner.

\section{Limitations}

While the Lang2MLIP shows promising results, several limitations remain.
First, our evaluation focuses on a single SEI system. 
Although complex, broader experiments are needed to assess generality.
Second, direct comparisons with fully optimized fixed pipelines or expert-designed workflows remain challenging. 
This is because constructing strong baselines is non-trivial, as both performance and required effort~(e.g., time, computational resources, and human expertise) depend sensitively on choices such as sampling strategy, curriculum design, and simulation conditions. Well-tuned pipelines can achieve strong performance, but often require substantial expert involvement and iterative tuning, making fair and reproducible comparisons challenging.
Our ablations partially address this gap: Ablation A approximates a minimal manual pipeline, while Ablation B corresponds to a fixed schedule without closed-loop adaptation. Both show degraded performance due to limited initialization or lack of iterative refinement.
We emphasize that the goal of this work is not to outperform expert-designed workflows, but to enable non-experts to obtain usable MLIPs through an automated, language-driven process. We leave systematic benchmarking against optimized baselines for future work. 
\section{Conclusion}

We presented Lang2MLIP, a language-driven multi-agent framework for end-to-end MLIP development from natural-language input.
On a challenging multilayer SEI system, the agent constructs a structured training process that progresses from simple components to interfaces and full systems, while incorporating failure-driven data acquisition and iterative model refinement.
Notably, this behavior emerges without an explicitly designed curriculum or manually specified workflow.
By autonomously organizing the training process and adapting both data selection and sampling strategies, the system enables users without extensive domain expertise to carry out complex MLIP development that would traditionally require careful manual design and multiple rounds of tuning.
More broadly, they suggest a shift toward language-
driven automation of scientific pipelines.

\section*{Impact Statement}

This work lowers the barrier to MLIP development by enabling end-to-end workflows from natural language. By automating decisions that typically require domain expertise, it broadens access to atomistic modeling and may accelerate research in areas such as energy and materials discovery. Even when expert validation is required, the system removes the need to manually design and tune active learning pipelines, significantly reducing development time.
However, increased automation may introduce risks. Non-experts may produce MLIPs that appear valid in limited evaluations but fail in broader applications, potentially leading to incorrect conclusions. In such cases, careful verification by people who know the applications remains important. At the same time, the system can substantially reduce development time by eliminating the need to manually design and tune complex active learning pipelines.
More broadly, this work reflects a shift toward language-driven automation of scientific workflows. If developed responsibly, such systems could make advanced computational methods more accessible while preserving reliability through appropriate safeguards and evaluation standards.







\bibliography{reference}
\bibliographystyle{icml2026}

\newpage
\appendix
\onecolumn
\appendix

\section{Lang2MLIP pseudocode}
Lang2MLIP execution steps are outlined in Algorithm~\ref{alg:lang2mlip}.
\begin{algorithm}
\caption{Lang2MLIP}
\label{alg:lang2mlip}
\begin{algorithmic}[1]
\STATE \textbf{Input:} Natural-language task description $x$
\STATE \textbf{Output:} Trained MLIP model $\mathcal{M}$

\STATE \textbf{Phase 1: Interactive preparation}
\STATE Initialize task context $\mathcal{T}$ from $x$
\WHILE{the task is not sufficiently specified}
    \STATE Query user for missing or task-specific information
    \STATE Update $\mathcal{T}$
\ENDWHILE
\STATE Use initial structure agents to generate initial structures 
\STATE (Optional) Prepare a reference MD script

\STATE \textbf{Phase 2: Autonomous training}
\STATE Initialize dataset $\mathcal{D}_0$, model $\mathcal{M}_0$, evaluation record $\mathcal{E}_0$, and execution log $\mathcal{L}_0$
\STATE $t \gets 0$
\WHILE{true}
    \STATE Form the state $s_t = (\mathcal{D}_t, \mathcal{M}_t, \mathcal{E}_t, \mathcal{L}_t)$
    \STATE The decision-making agent selects an action $a_t \in \mathcal{A}$
    \IF{$a_t = \texttt{End}$}
        \STATE \textbf{break}
    \ENDIF
    \STATE The decision-making agent invokes the corresponding action agent for $a_t$
    \STATE Execute $a_t$ and obtain updated outputs
    \STATE Update $\mathcal{D}_{t+1}$, $\mathcal{M}_{t+1}$, $\mathcal{E}_{t+1}$, and $\mathcal{L}_{t+1}$
    \STATE $t \gets t + 1$
\ENDWHILE

\STATE \textbf{return} $\mathcal{M}_t$
\end{algorithmic}
\end{algorithm}

\clearpage

\section{Validation and ablation test results}
\label{app:val_ablation}
\subsection{Density}

This subsection compares the time evolution of the simulation-box density predicted by PFP, the Lang2MLIP model, and the ablation model without autonomous iterative refinement.

\begin{figure}[H]
  \centering
  \includegraphics[width=0.5\columnwidth]{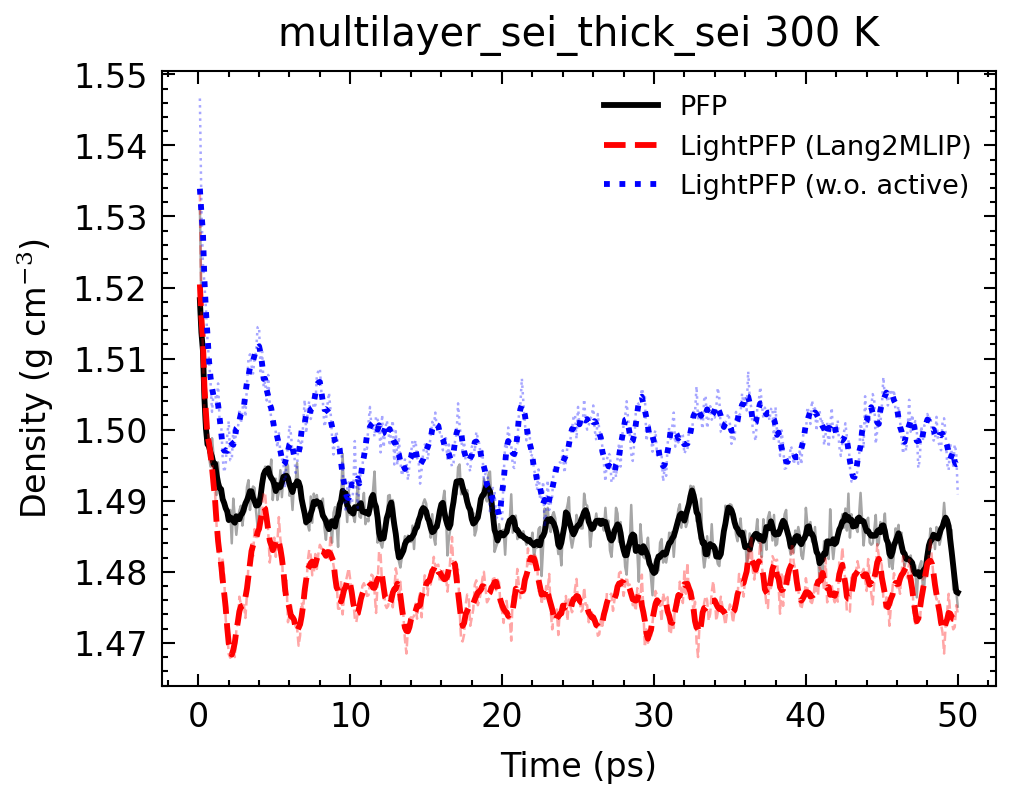} %
  \caption{Time evolution of the simulation-box density for the multilayer SEI system at 300 K.}
  \label{fig:ablation_density}
\end{figure}

\subsection{Mean squared displacement (MSD)}

This subsection presents the mean squared displacement curves for different elements in the SEI simulation box, providing a dynamical comparison between PFP, full Lang2MLIP model, and the ablation model.

\begin{figure}[H]
  \centering
  \includegraphics[width=0.8\columnwidth]{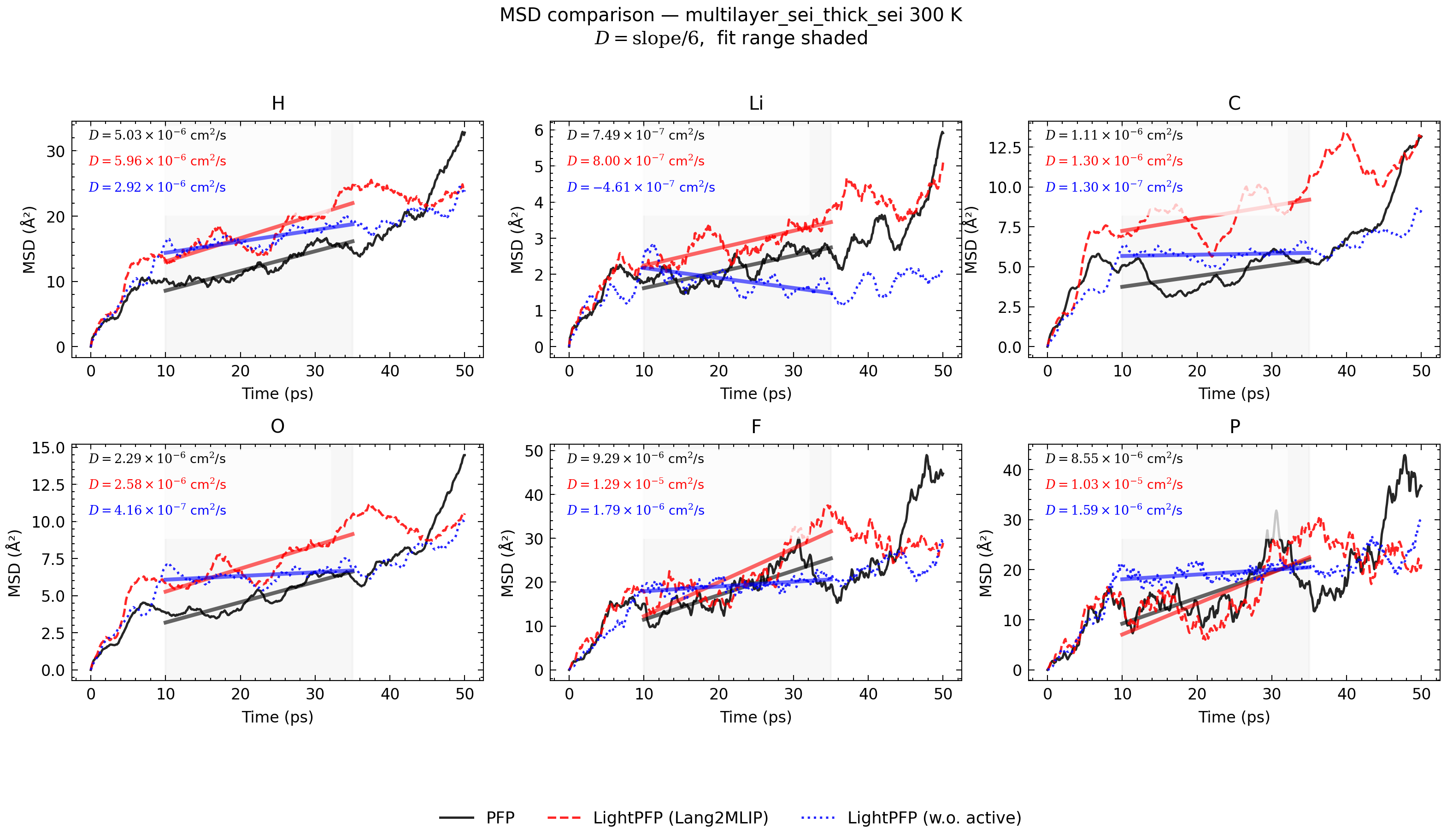} %
  \caption{Mean squared displacement (MSD) curves of all elements in the multilayer SEI simulation box at 300 K.}
  \label{fig:ablation_msd}
\end{figure}

\subsection{Radial distribution function}

This subsection reports RDF comparisons for different regions of the SEI system, including the anode, inorganic SEI, organic SEI, and electrolyte, in order to evaluate how well the trained models reproduce local atomic structure.

\begin{figure}[H]
  \centering
  \includegraphics[width=0.6\columnwidth]{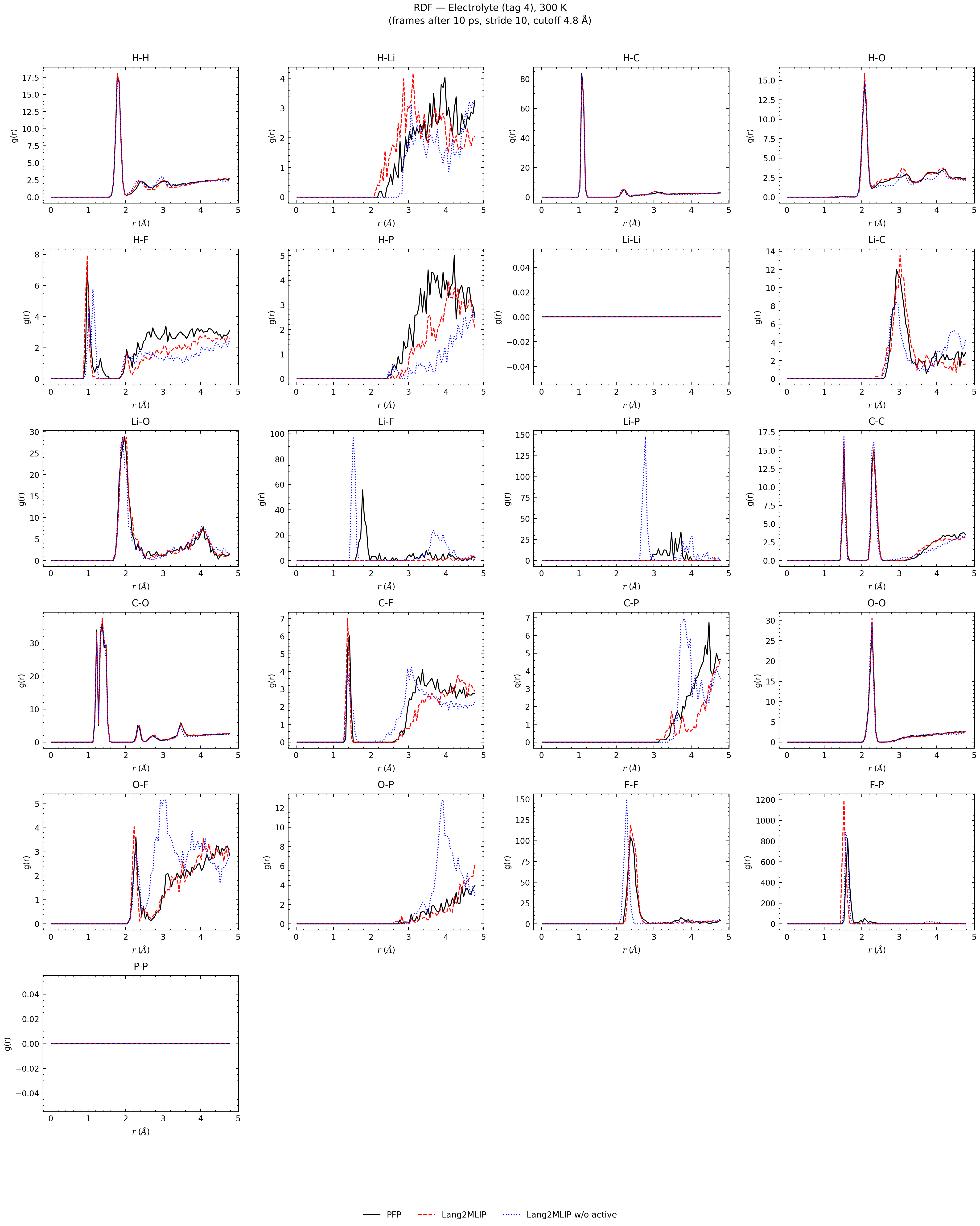} %
  \caption{Radial distribution functions (RDFs) of the electrolyte (LiPF6/EC/DMC) region in the multilayer SEI system at 300 K}
  \label{fig:ablation_electrolyte_rdf}
\end{figure}

Some unusual features are observed in the electrolyte RDF (Figure ~\ref{fig:ablation_electrolyte_rdf}) due to the low concentration of LiPF$_6$ in the EC/DMC electrolyte, which leads to poor sampling statistics for pair correlations containing Li and P. The nearly flat zero profiles for Li--Li and P--P are physically reasonable in this setting. For Li--Li, Li$^+$ is a dilute cation and two Li$^+$ ions are unlikely to appear within the short RDF cutoff because electrostatic repulsion and solvent screening generally keep them apart. For P--P, each P atom is the center of a PF$_6^-$ anion, and since PF$_6^-$ is also dilute, close P--P pair occurrences are likewise extremely rare within the sampled configurations. As a result, both Li--Li and P--P correlations can remain essentially zero over the plotted distance range without implying any unphysical behavior. By contrast, the larger discrepancies among the three curves for Li--F and Li--P likely reflect the sensitivity of these correlations to the local ion-pairing and dissociation states of LiPF$_6$. In a dilute electrolyte, Li$^+$ ions may alternate between solvent-separated and contact-associated environments with PF$_6^-$, so the corresponding Li--F and Li--P distributions are more weakly defined and more sensitive to finite-size effects, limited trajectory length, and model-dependent differences in local solvation structure. Therefore, rather than indicating the absence of any underlying physical structure, these results more likely suggest that the Li/P-related RDFs are statistically less well converged and intrinsically more variable than the dominant solvent-related correlations in the electrolyte region.

\begin{figure}[H]
  \centering
  \includegraphics[width=0.4\columnwidth]{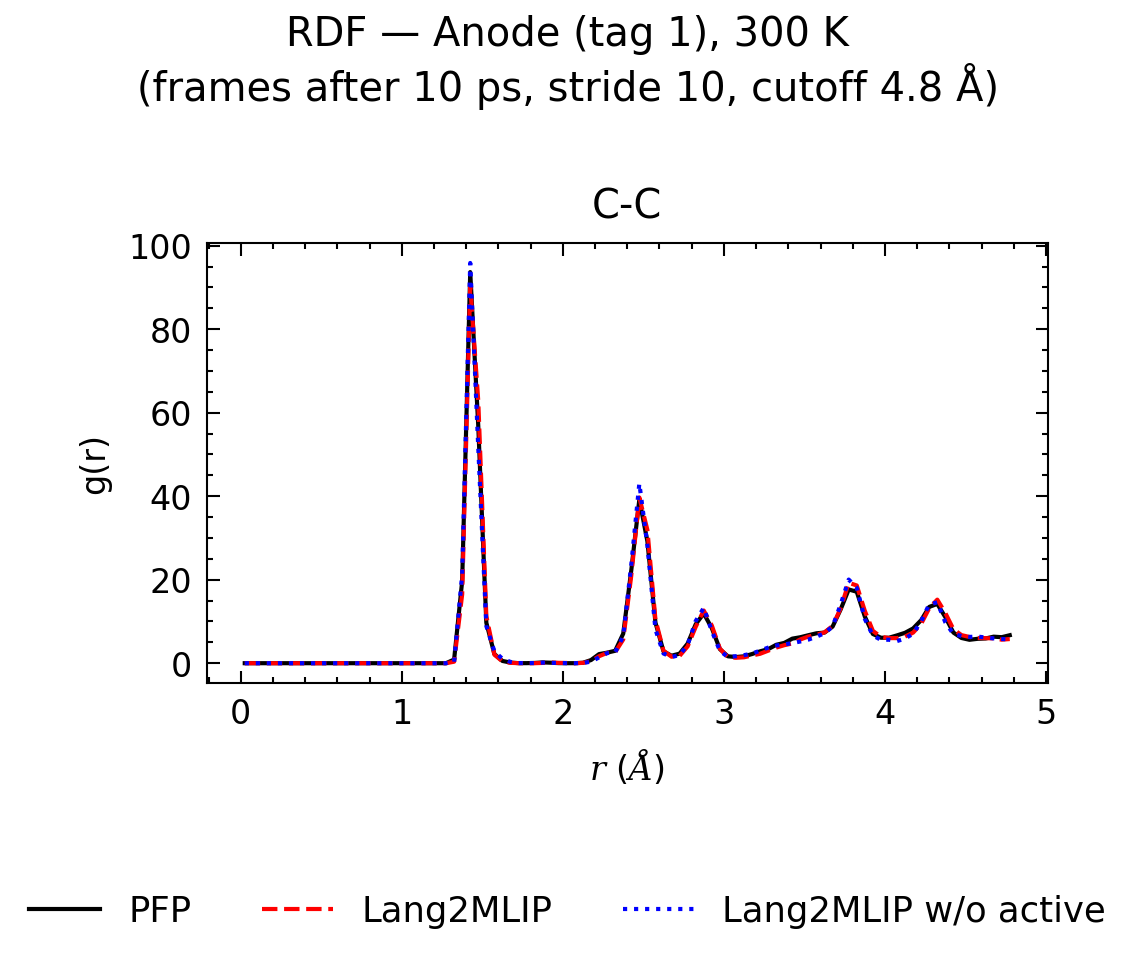} %
  \caption{Radial distribution functions (RDFs) of the anode region (graphite) in the multilayer SEI system at 300 K.}
  \label{fig:ablation_anode_rdf}
\end{figure}

\begin{figure}[H]
  \centering
  \includegraphics[width=0.6\columnwidth]{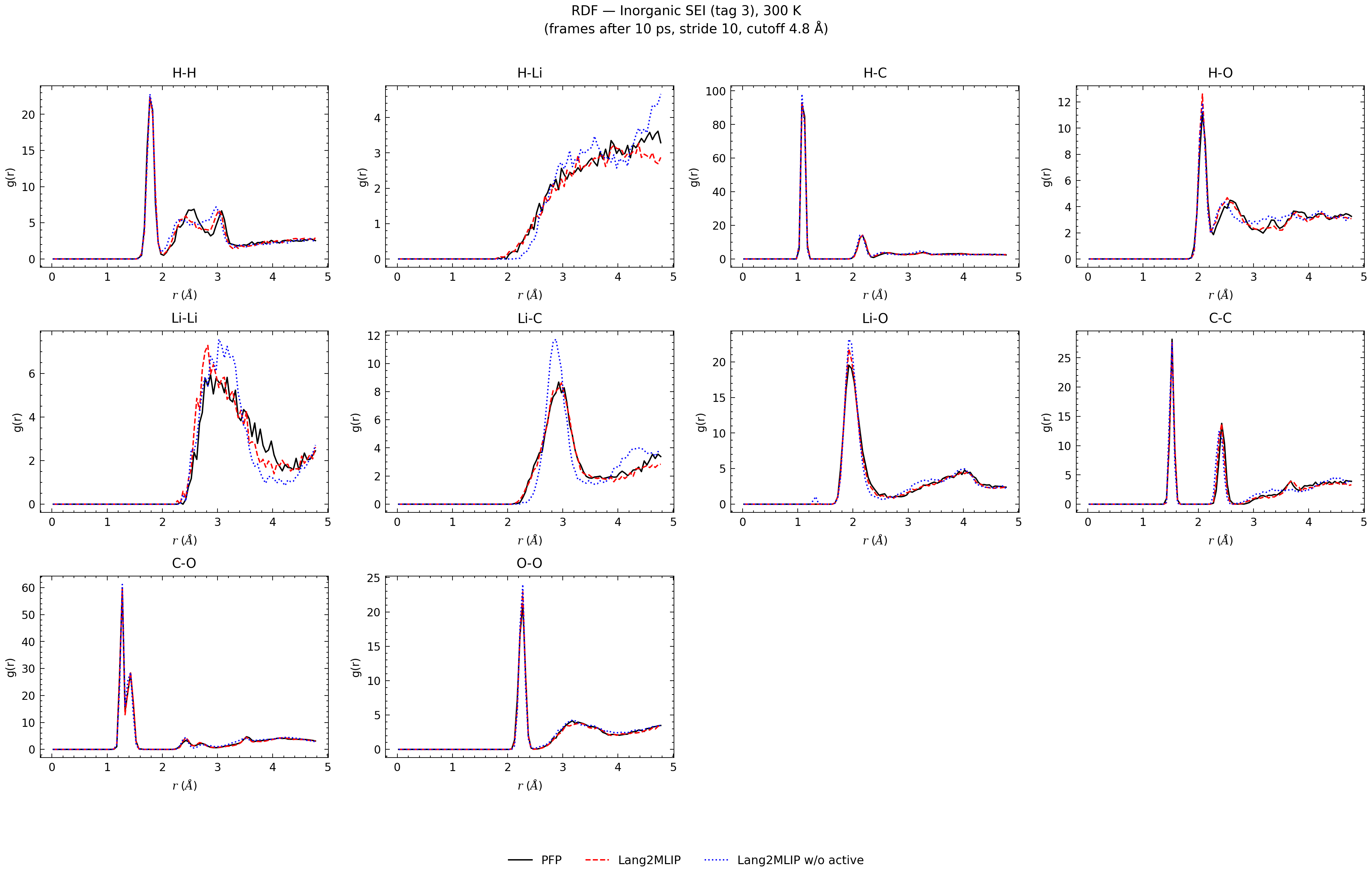} %
  \caption{Radial distribution functions (RDFs) of the inorganic SEI (Li2CO3) region in the multilayer SEI system at 300 K}
  \label{fig:ablation_inorganic_sei_rdf}
\end{figure}

\begin{figure}[H]
  \centering
  \includegraphics[width=0.6\columnwidth]{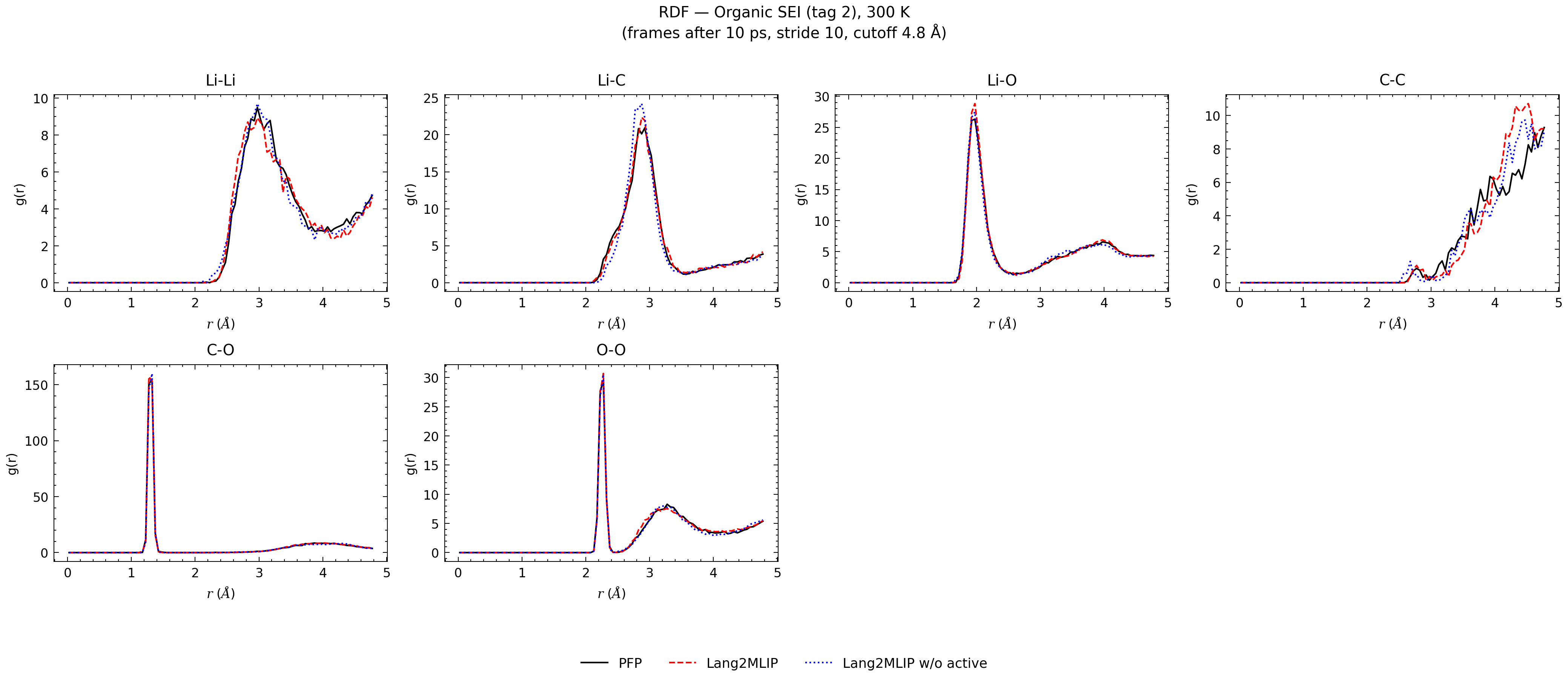} %
  \caption{Radial distribution functions (RDFs) of the organic SEI (LEDC) region in the multilayer SEI system at 300 K}
  \label{fig:ablation_organic_sei_rdf}
\end{figure}

\clearpage

\section{Initial structures}
\label{app:init_structures}

This appendix presents 60 initial structures related to SEI generated by the preparation agents.

\begin{figure}[H]
    \centering
    \includegraphics[width=0.65\textwidth]{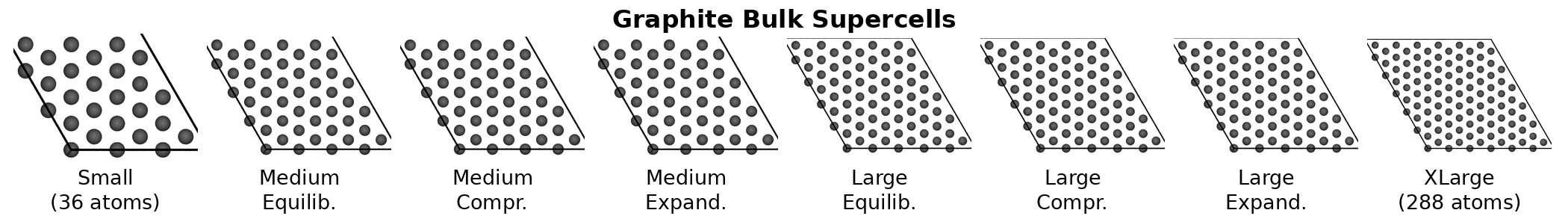}
    \vspace{0.5cm}
    \includegraphics[width=0.65\textwidth]{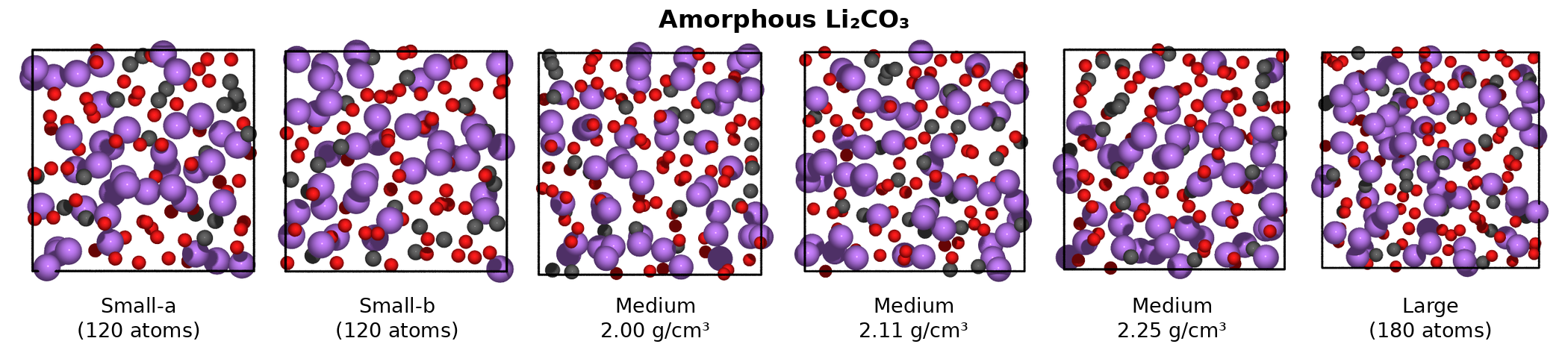}
    \vspace{0.5cm}
    \includegraphics[width=0.65\textwidth]{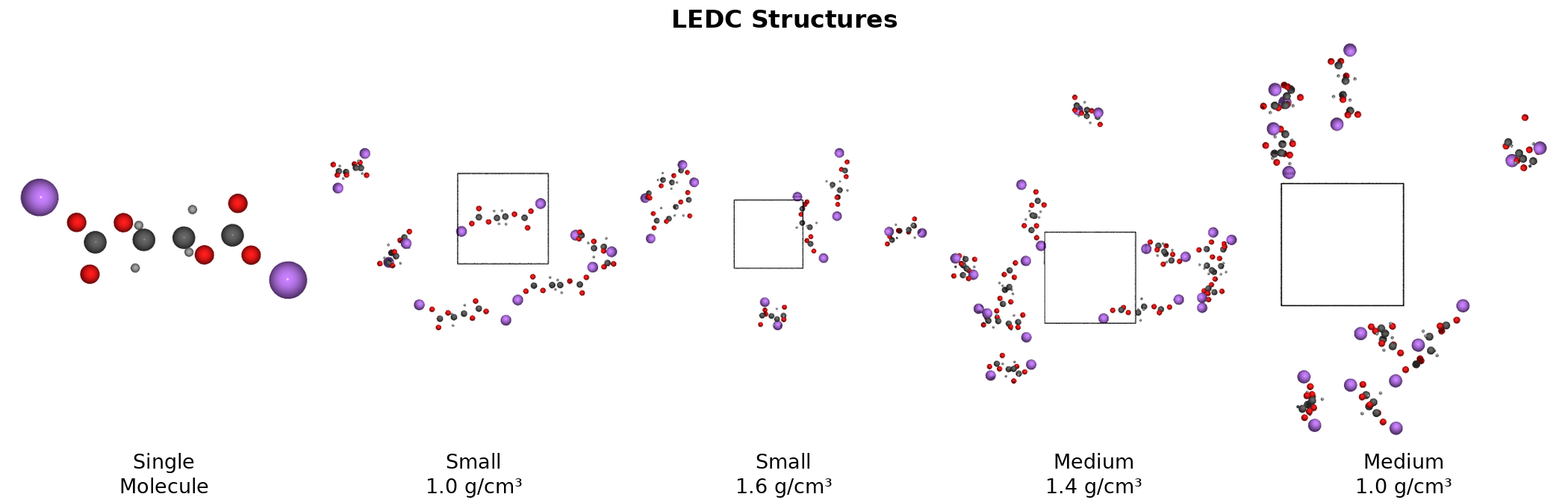}
    \vspace{0.5cm}
    \includegraphics[width=0.65\textwidth]{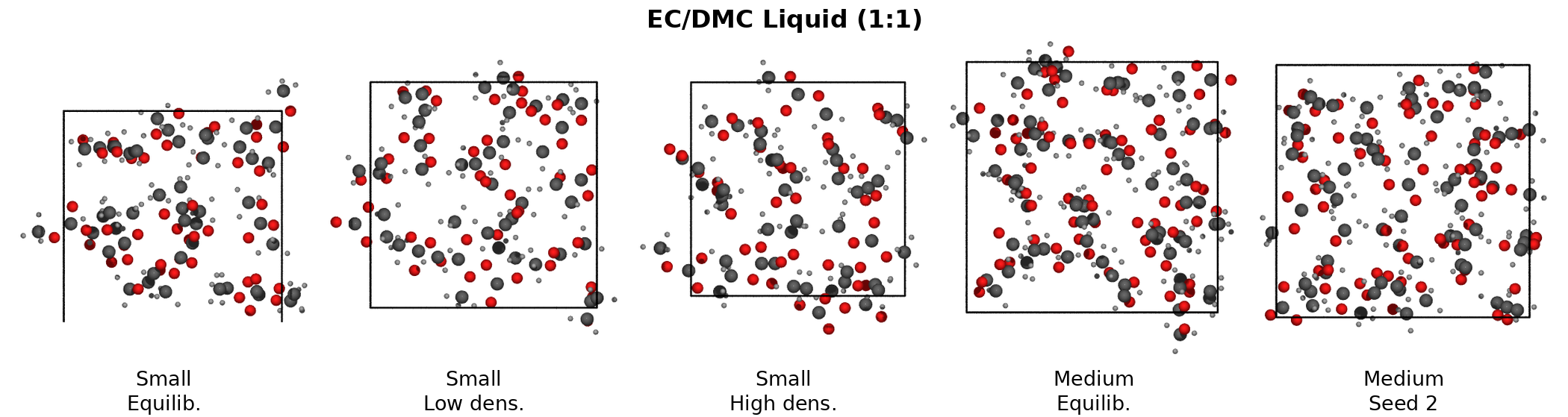}
    \vspace{0.5cm}
    \includegraphics[width=0.65\textwidth]{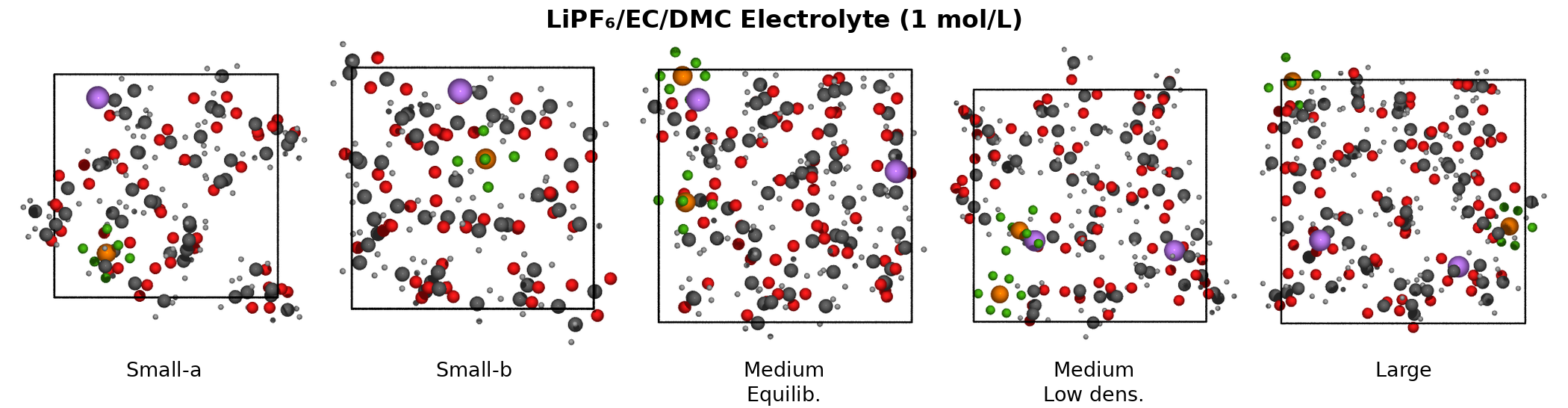}
    \vspace{0.5cm}
    \includegraphics[width=0.65\textwidth]{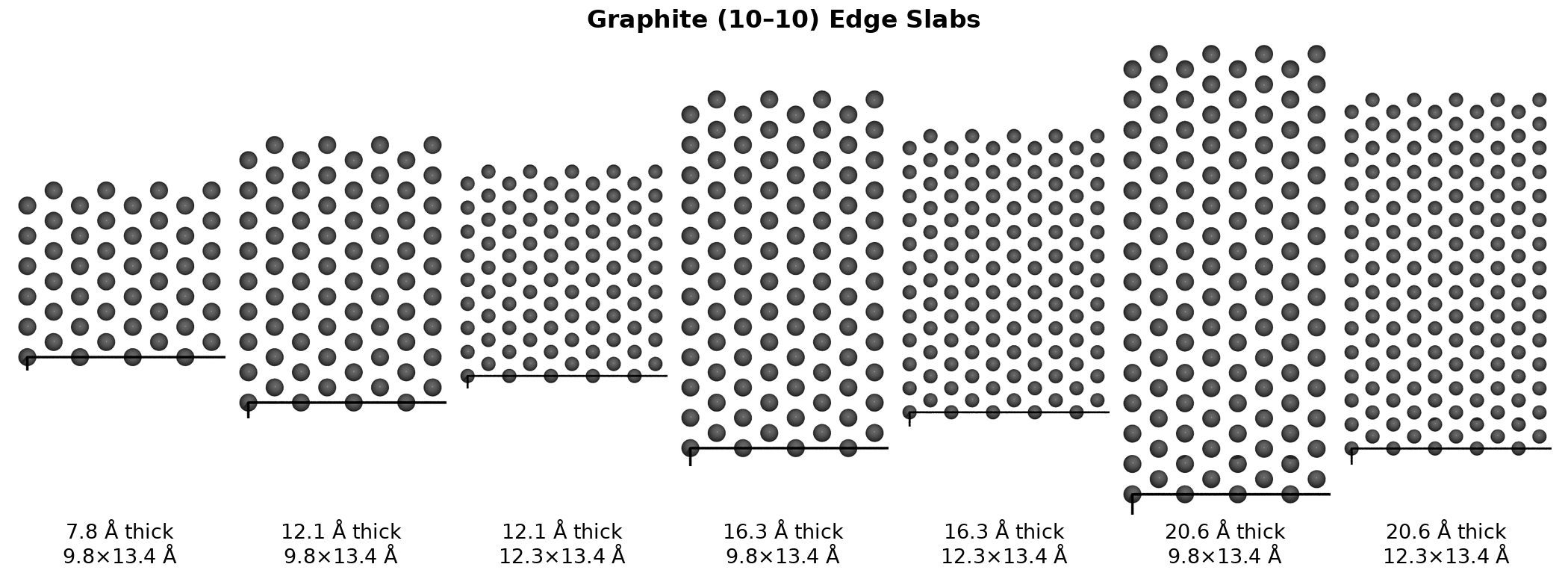}
    \caption{Initial structures of basic components of SEI generated by preparation agents}
    \label{fig:init_structure_basic}
\end{figure}

\begin{figure}[h]
    \centering
    \includegraphics[width=0.8\textwidth]{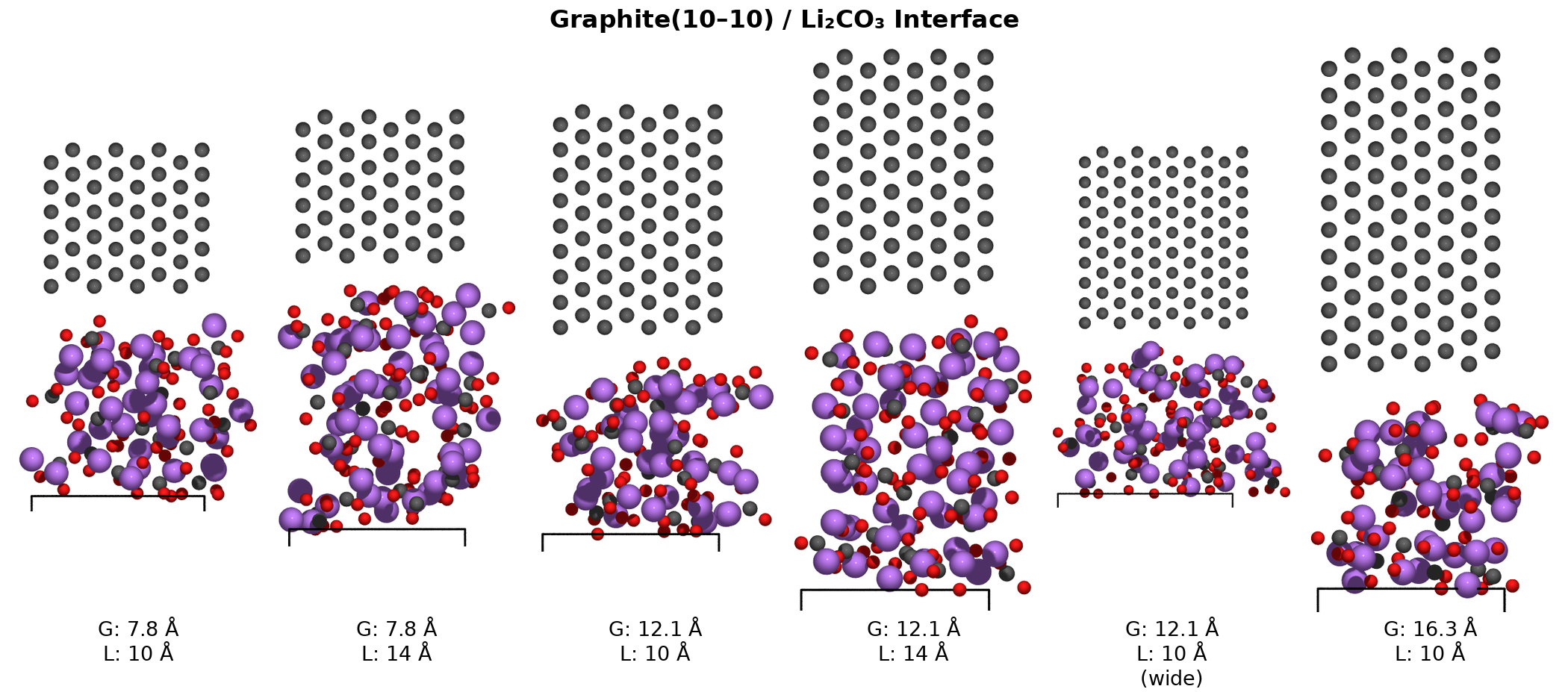}
    \vspace{0.5cm}
    \includegraphics[width=0.8\textwidth]{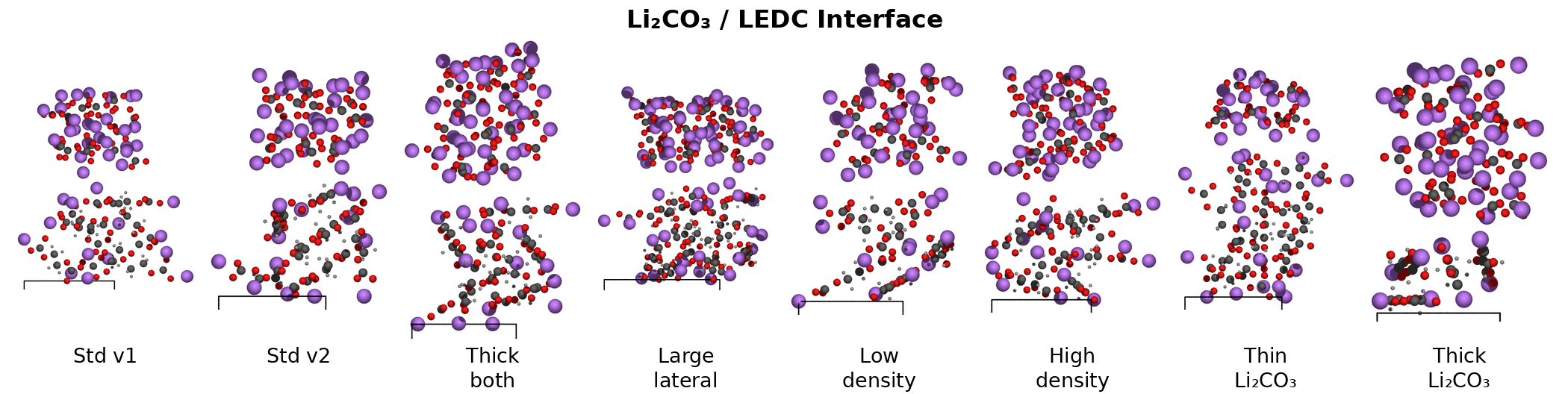}
    \vspace{0.5cm}
    \includegraphics[width=0.8\textwidth]{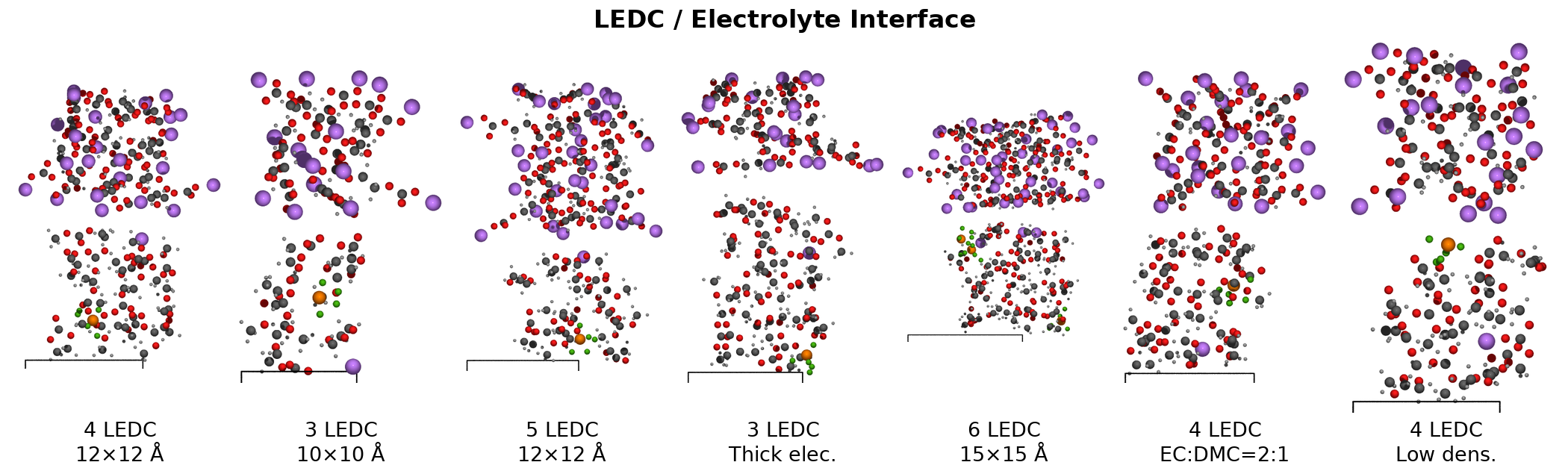}
    \caption{Initial structures of interfaces in SEI generated by preparation agents}
    \label{fig:init_structure_interface}
\end{figure}

\begin{figure*}
  \centering
  \includegraphics[width=0.5\columnwidth]{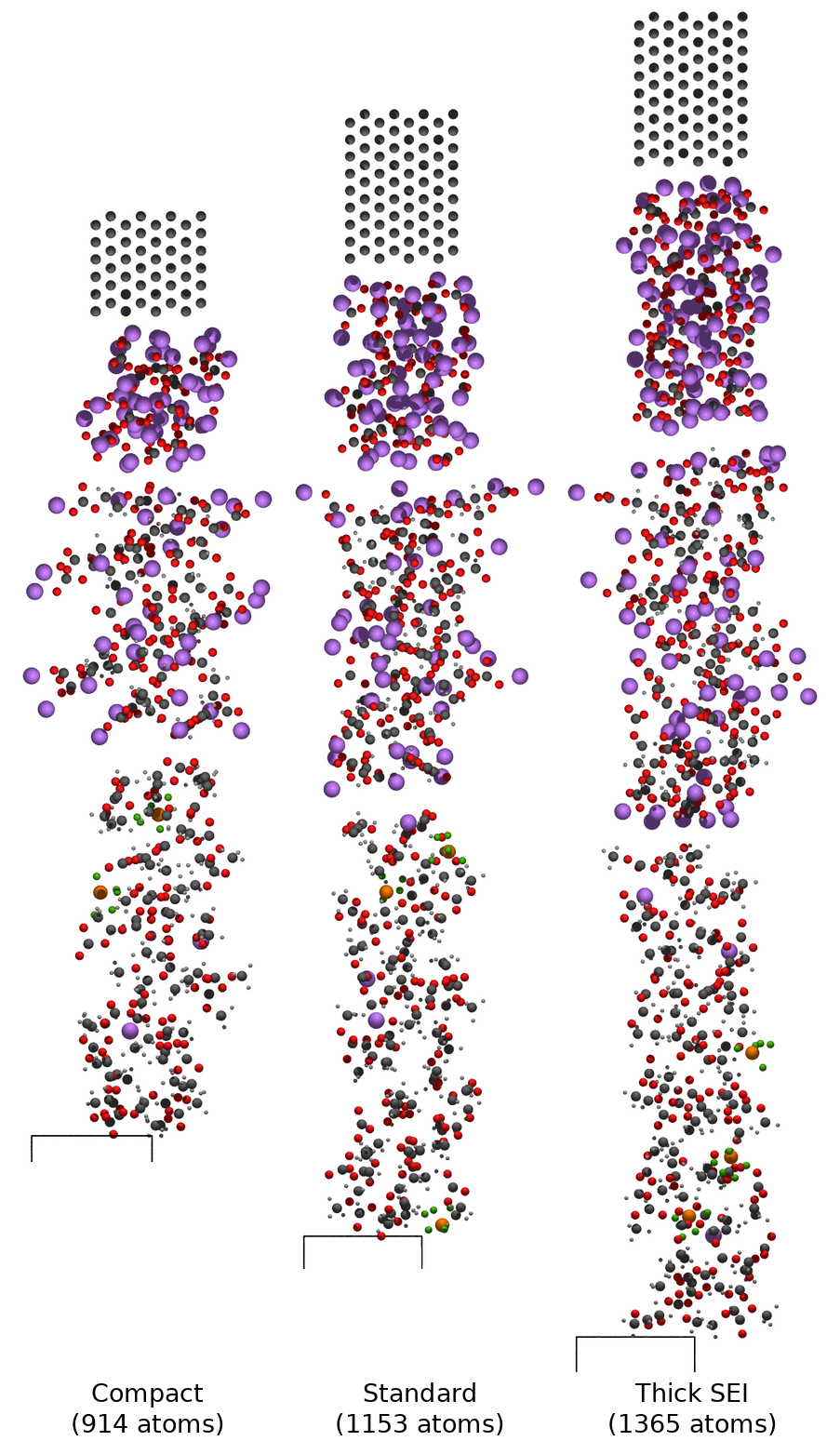} %
  \caption{Initial structures of SEI multiple layers structures generated by preparation agents}
  \label{fig:init_structure_multilayers}
\end{figure*}

\clearpage

\section{Information Collection Log}
\label{app:input_prompt}

This appendix provides the information collection log from the interactive preparation phase, showing how the agent progressively clarified the user’s task through dialogue.

\begin{flushleft}
\small

Step 1: Gathering information...\\
\noindent\rule{\textwidth}{0.4pt}\\
Information Collection Started.\\
\noindent\rule{\textwidth}{0.4pt}

\vspace{0.5em}
Please input the description of simulation task...

\vspace{0.5em}
\textbf{User}: SEI of lithum battery

\vspace{0.5em}
\textbf{Claude}: Hello! The simulation of SEI (Solid Electrolyte Interphase) is a very interesting and important research topic. The SEI layer is crucial to the performance and lifetime of lithium batteries.

To better help you prepare MLIP training, I would like to first understand some more specific information.

First, the composition and structure of SEI can be very complex---it usually contains organic components (such as Li$_2$CO$_3$, ROCO$_2$Li, etc.) and inorganic components (such as LiF, Li$_2$O, etc.), and is closely related to the choice of electrolyte.

Could you please tell me:
\begin{itemize}
    \item Which \textbf{electrolyte system} are you focusing on (e.g., EC/DMC-based, ether-based electrolyte, or solid-state electrolyte)?
    \item Do you want to simulate the \textbf{formation process} of SEI, or certain \textbf{properties} of an already formed SEI layer (such as ionic conductivity, mechanical properties, etc.)?
\end{itemize}

This will help me better understand your research goals.

\noindent\rule{\textwidth}{0.4pt}\\
Collection Status:\\
\noindent\rule{\textwidth}{0.4pt}\\
Completed: False\\
Summary:\\
Next Question: Which electrolyte system are you focusing on (e.g., EC/DMC-based, ether-based electrolyte, or solid-state electrolyte)? Do you want to simulate the formation process of SEI, or certain properties of an already formed SEI layer (such as ionic conductivity, mechanical properties, etc.)?\\
\noindent\rule{\textwidth}{0.4pt}

\vspace{0.5em}
\textbf{User}: Electrolyte contains EC/DMC and LiPF6. I want to get the structure of SEI interface.

\vspace{0.5em}
\textbf{Claude}: Got it! EC/DMC + LiPF6 is the most classic lithium battery electrolyte system, and studying the structure of SEI is very valuable.

Next, I would like to understand a few more aspects:
\begin{enumerate}
    \item \textbf{Anode material} --- Are you using a graphite anode? Or other materials (such as silicon-based anodes, lithium metal anodes)? This will affect the construction of the SEI interface structure.
    \item \textbf{Simulation method} --- In order to obtain the interface structure of SEI, what simulation strategy do you plan to use? For example:
    \begin{itemize}
        \item Growing SEI by simulating the decomposition reactions of electrolyte on the anode surface using molecular dynamics (MD)?
        \item Or constructing an SEI model with known composition, followed by structural relaxation/annealing?
    \end{itemize}
    \item \textbf{PFP version} --- Which version of PFP do you plan to use? And which calc mode (e.g., \texttt{crystal}, \texttt{crystal\_plus\_d3}, etc.)?
\end{enumerate}

\noindent\rule{\textwidth}{0.4pt}\\
Collection Status:\\
\noindent\rule{\textwidth}{0.4pt}\\
Completed: False\\
Summary:\\
Next Question: Is the anode material graphite or something else? What simulation method do you plan to use to obtain the SEI structure (e.g., MD simulation of decomposition process, or relaxation after constructing a known SEI model)? And which PFP version and calc mode do you plan to use?\\
\noindent\rule{\textwidth}{0.4pt}

\vspace{0.5em}
\textbf{User}: Anode is graphite. I want to build the structure of know SEI structure, which contains inorganic layer (amorphous Li2CO3) and organic layer (LEDC). PFP version is v8.0.0 and calc mode is PBE

\vspace{0.5em}
\textbf{Claude}: Very good, the information is becoming clearer! This is a classic layered SEI model---on the graphite anode, there is first an inorganic layer (amorphous Li$_2$CO$_3$), followed by an organic layer (LEDC, i.e., (CH$_2$OCO$_2$Li)$_2$).

To refine the simulation plan, I would like to confirm a few more points:
\begin{enumerate}
    \item \textbf{Construction of the interface model} --- Is the structure you envision arranged in the following layered order?
    \begin{itemize}
        \item Graphite (bottom) $\rightarrow$ amorphous Li$_2$CO$_3$ (inorganic layer) $\rightarrow$ LEDC (organic layer) $\rightarrow$ electrolyte (top)
        \item Or are you only focusing on the SEI itself (Li$_2$CO$_3$ + LEDC), without including graphite and electrolyte?
    \end{itemize}
    \item \textbf{Simulation strategy} --- After constructing the initial structure, what method do you plan to use to obtain a reasonable SEI structure? For example:
    \begin{itemize}
        \item High-temperature annealing MD $\rightarrow$ cooling relaxation?
        \item Direct structural optimization?
        \item Or melt-quench to generate amorphous Li$_2$CO$_3$?
    \end{itemize}
    \item \textbf{Temperature and system size} --- Do you have preferences for target temperature and system size?
\end{enumerate}

\noindent\rule{\textwidth}{0.4pt}\\
Collection Status:\\
\noindent\rule{\textwidth}{0.4pt}\\
Completed: False\\
Summary:\\
Next Question: Please confirm: (1) Does the interface model include graphite and electrolyte, or only focus on the SEI itself? (2) What simulation strategy do you plan to use to obtain a reasonable structure (e.g., annealing MD, structural optimization, melt-quench, etc.)? (3) Do you have preferences for temperature and system size?\\
\noindent\rule{\textwidth}{0.4pt}

\vspace{0.5em}
\textbf{User}: The whole system contains graphite / inorganic SEI / organic SEI and electrolyte. I will get structure from annealing. I will simulate at round 300K.

\vspace{0.5em}
\textbf{Claude}: Thank you for providing the detailed information! I now have a relatively complete understanding of your simulation requirements. Let me summarize for confirmation:

\vspace{0.5em}
\noindent\textbf{Simulation system: SEI interface structure of lithium battery}
\begin{itemize}
    \item \textbf{PFP version}: v8.0.0, calc mode: PBE
    \item \textbf{System composition (from bottom to top)}:
    \begin{itemize}
        \item \textbf{Graphite anode} (bottom layer)
        \item \textbf{Inorganic SEI layer}: amorphous Li$_2$CO$_3$
        \item \textbf{Organic SEI layer}: LEDC ((CH$_2$OCO$_2$Li)$_2$)
        \item \textbf{Electrolyte}: EC/DMC + LiPF6 (top layer)
    \end{itemize}
    \item \textbf{Simulation objective}: Obtain a reasonable structure of the SEI interface
    \item \textbf{Simulation method}: Molecular dynamics (MD) annealing simulation
    \item \textbf{Target temperature}: $\sim$300K
    \item \textbf{Simulation workflow}: First construct a layered initial structure, then obtain a reasonable SEI interface structure through annealing MD
\end{itemize}

Is this summary accurate? Is there anything that needs to be added or modified?

\noindent\rule{\textwidth}{0.4pt}\\
Collection Status:\\
\noindent\rule{\textwidth}{0.4pt}\\
Completed: False\\
Summary:\\
Next Question: Is the above summary accurate? Is there anything that needs to be added or modified?\\
\noindent\rule{\textwidth}{0.4pt}

\vspace{0.5em}
\textbf{User}: The molar ratio between EC/DMC is 1:1. The LiPF6 concentation is 1 mol/L. The surface of graphite I intrest is graphite(10-10)

\vspace{0.5em}
\textbf{Claude}: Got it, the orientation of the graphite surface is now clear! Let me provide the final summary for confirmation:

\vspace{0.5em}
\noindent\textbf{Simulation system: SEI interface structure of lithium battery}
\begin{itemize}
    \item \textbf{PFP version}: v8.0.0, calc mode: PBE
    \item \textbf{System composition (from bottom to top)}:
    \begin{itemize}
        \item \textbf{Graphite anode} (bottom layer): graphite (10-10) surface (i.e., edge surface)
        \item \textbf{Inorganic SEI layer}: amorphous Li$_2$CO$_3$
        \item \textbf{Organic SEI layer}: LEDC ((CH$_2$OCO$_2$Li)$_2$)
        \item \textbf{Electrolyte} (top layer):
        \begin{itemize}
            \item Solvent: EC/DMC, molar ratio 1:1
            \item Lithium salt: LiPF$_6$, concentration 1 mol/L
        \end{itemize}
    \end{itemize}
    \item \textbf{Simulation objective}: Obtain a reasonable structure of the SEI interface
    \item \textbf{Simulation method}: Molecular dynamics (MD) annealing simulation
    \item \textbf{Target temperature}: $\sim$300K
    \item \textbf{Simulation workflow}: First construct a layered initial structure, then obtain a reasonable SEI interface structure through annealing MD
\end{itemize}

Is this final summary complete and accurate? Once confirmed, I will proceed to prepare the information required for MLIP training based on this.

\noindent\rule{\textwidth}{0.4pt}\\
Collection Status:\\
\noindent\rule{\textwidth}{0.4pt}\\
Completed: False\\
Summary:\\
Next Question: Is the final summary complete and accurate? After confirmation, I will start preparing the information required for MLIP training.\\
\noindent\rule{\textwidth}{0.4pt}

\vspace{0.5em}
\textbf{User}: Yes, it is correct

\vspace{0.5em}
\textbf{Claude}: Great! Thank you for your confirmation. The information collection is now complete.
Information summary saved to \texttt{information\_summary\_report.txt}

\end{flushleft}

\clearpage

\section{Prompt of Decision Maker for MLIP Active Learning}
\label{app:decision-maker}

This appendix contains the full prompt for the decision-making agent used in the autonomous training phase.

\subsection*{Role}
You are a top-level decision maker, responsible for determining and outputting the ``next task to execute'' and the corresponding specific parameters/settings, based on the existing dialogue records and structured reports, in order to advance the MLIP (machine-learned interatomic potential) active learning workflow.  
You do not directly run execution tasks such as DFT/MD/training; instead, based on the current information and strategy, you explicitly assign the next task and provide sufficiently precise parameters so that the executor (or subordinate agent) can execute it immediately.

\subsection*{Key Points}
\begin{itemize}
\item When the task is complex, first carry out basic training on simple basic units and structures, and gradually transition to the complex complete target material. For example, for an interface task involving SiO$_2$ and Al$_2$O$_3$, training and validation can first be performed on the pure bulk structures of SiO$_2$ and Al$_2$O$_3$, and only afterward transition to training and validation on interface structures.
\item The decision should control the overall direction of task execution, without restricting task details.
\item Accuracy targets should be tied to task relevance: structures directly related to the target property (for example, interface structures when calculating interface diffusion) require sufficient sampling and high-accuracy coverage; indirectly related components (for example, pure bulk phases) only need to reach basic accuracy, and do not require excessive resource investment.
\end{itemize}

\subsection*{Available Information}

At each decision step, the Agent can access the following content:

\begin{itemize}
\item \texttt{information\_summary\_report.txt} --- overall task description (the user-provided target, the purpose of using MLIP for the target, the physical properties expected to be calculated, etc.)
\item \texttt{dialogue.log} --- previous dialogue records, including the communication content between the decision maker and the execution agents
\item \texttt{init\_structures/init\_structure\_description.txt} --- textual descriptions / annotations for each initial structure
\item \texttt{reference\_md.txt} --- (if present) description of unusual MD simulation methods involved in the simulation task
\end{itemize}

\subsection*{Goals}

Based on the context (task description + dialogue records + structure reports + files) and the available information, choose the most reasonable next task. The task can only be one of the following options.

\begin{itemize}
\item \texttt{eval\_reference} (only at the very beginning of the task, run MD with PFP on structures labeled as validation, save the trajectory, and collect basic physical information)
\item \texttt{pfp\_sample} (only at the very beginning of the task, use PFP to comprehensively sample all initial structures and generate the initial training dataset)
\item \texttt{sample} (run MD sampling using the existing MLIP or PFP)
\item \texttt{selection} (select samples and submit them for reference calculations)
\item \texttt{train} (train or fine-tune the MLIP)
\item \texttt{evaluation} (model evaluation)
\item \texttt{prune} (delete data / outliers / MLIPs)
\item \texttt{end} (terminate the entire active learning task and provide final recommendations)
\end{itemize}

\subsection*{Output Requirements}

Decision output format (must be followed strictly for downstream parsing)

JSON Schema (example)
\begin{verbatim}
{
  "next_task": "string",          // optional values: "pfp_sample", "eval_reference", "sample", "selection", "train", "evaluation", "prune", "end"
  "descriptions": "string"        // short human-readable explanation of why this task was chosen
}
\end{verbatim}

\subsection*{\texttt{eval\_reference}}

\subsubsection*{What is the \texttt{eval\_reference} task}
The \texttt{eval\_reference} task uses the PFP calculator to run MD simulations on structures in the initial structures that are labeled for validation use, and saves the trajectories.

This task has two purposes:
\begin{itemize}
\item To collect the basic physical information of these structures (density, interatomic distances, potential energy, etc.) for reference by the \texttt{sample} task.
\item To generate validation-specific trajectories for direct use in subsequent \texttt{evaluation} tasks (the \texttt{evaluation} task only needs to run LightPFP calculations on these trajectories).
\end{itemize}

\subsubsection*{When to choose \texttt{eval\_reference}}
\begin{itemize}
\item Run only once at the very beginning of the task.
\end{itemize}

\subsubsection*{What information the task description should include}
\begin{itemize}
\item Only indicate that the \texttt{eval\_reference} task has been selected.
\end{itemize}

\subsection*{\texttt{pfp\_sample}}

\subsubsection*{What is the \texttt{pfp\_sample} task}
The \texttt{pfp\_sample} task uses the PFP calculator to comprehensively sample all initial structures (except those used for validation), generating a high-quality HDF5 initial training dataset for use in the first \texttt{train} task.

\subsubsection*{When to choose \texttt{pfp\_sample}}
\begin{itemize}
\item Run only once at the very beginning of the task.
\end{itemize}

\subsubsection*{What information the task description should include}
Only need to indicate that the \texttt{pfp\_sample} task has been selected.

\subsubsection*{Key Point}
After \texttt{pfp\_sample} is completed, the next step should go directly to \texttt{train}, with no additional \texttt{selection} required.

\subsection*{\texttt{sample}}

\subsubsection*{What is the \texttt{sample} task}
The \texttt{sample} task uses an existing machine-learned interatomic potential (MLIP) or a reference method (such as PFP) to run molecular dynamics (MD) simulations, in order to generate new atomic structure samples for expanding the training dataset and improving the performance and generalization capability of the MLIP.  
The \texttt{sample} task generates MD trajectory files for subsequent sample selection and reference calculations.

\subsubsection*{When to choose \texttt{sample}}
\begin{itemize}
\item The MLIP has not reached sufficient stability and accuracy, and more samples are needed to cover unseen structure space.
\end{itemize}

\subsubsection*{What information the task description should include}
\begin{itemize}
\item What kind of representation the goal is to learn or enhance, for example, the interaction between Si and O atoms
\item What type of initial structures are recommended to be selected (they must exist in \texttt{init\_structures/init\_structure\_description.txt}, without specifying a particular structure)
\item Whether to use the MLIP (with the model specified) or a reference method for sampling
\item The basic condition range of the sampling MD, for example, the temperature range
\end{itemize}

\subsubsection*{Key Points}
\begin{itemize}
\item Read the task description and understand the user's goal and the physical properties expected to be calculated.
\item Read \texttt{init\_structures/init\_structure\_description.txt} to understand the types of initial structures already available.
\item Read \texttt{dialogue.log} to understand the current task progress.
\item Ensure diversity in sampling, covering different structure types and MD conditions, especially those that have not been sampled before, in order to improve the model's generalization ability.
\item When an MLIP is available, give priority to using the MLIP for sampling.
\end{itemize}

\subsection*{\texttt{selection}}

\subsubsection*{What is the \texttt{selection} task}
The \texttt{selection} task involves selecting the most valuable samples from existing atomic structure samples and submitting them to a reference calculation method (such as DFT) for high-accuracy calculations. These samples will be used to expand and improve the training dataset of the machine-learned interatomic potential (MLIP).  
After the \texttt{selection} task is completed, a dataset file is usually generated for subsequent model training.

\subsubsection*{When to choose \texttt{selection}}
\begin{itemize}
\item In the initial stage of active learning, after using a reference method to run MD tasks, when the MD trajectory files need to be converted into a training dataset.
\item After the preceding MD sampling has generated a large number of new atomic structure samples, sufficient to select the most valuable samples from them for reference calculations.
\end{itemize}

\subsubsection*{What information the task description should include}
\begin{itemize}
\item One or more MD trajectory files as the source of the samples.
\item Whether all samples are to be selected, or only part of them.
\item If only part of them are to be selected, the recommended number of selected samples or sampling ratio must be clearly specified.
\end{itemize}

\subsubsection*{Key Points}
\begin{itemize}
\item If the MD trajectory files were obtained from MD tasks run with a reference method (such as PFP), it is usually recommended to select all of these samples for reference calculations.
\item If the MLIP model performed unstably during sampling, or if the number of samples selected in previous \texttt{selection} tasks was small, it is possible to recommend selecting more samples to accelerate model improvement, for example, 30\% of the samples. If the MLIP model is already relatively stable, or if a large number of samples were selected in previous \texttt{selection} tasks, it is possible to recommend selecting fewer samples to save computational resources, for example, 5--10\% of the samples.
\end{itemize}

\subsection*{\texttt{train}}

\subsubsection*{What is the \texttt{train} task}
The \texttt{train} task uses the existing training dataset to train the machine-learned interatomic potential (MLIP).  
After the \texttt{train} task is completed, a new MLIP model is generated for subsequent MD sampling and evaluation.

\subsubsection*{When to choose \texttt{train}}
\begin{itemize}
\item If a \texttt{selection} task has been carried out previously and a sufficient amount of new training dataset has been generated, then the \texttt{train} task can be chosen to train the MLIP.
\item If a \texttt{prune} task was carried out previously because of dataset issues, some outliers or inappropriate data were removed, and this is judged to have a significant impact on the model, retraining is required.
\item The previous model training task failed, or the performance and stability are very poor, and the model needs to be retrained.
\end{itemize}

\subsubsection*{What information the task description should include}
\begin{itemize}
\item Whether to train from scratch or continue training from an existing model
\item If continuing training from an existing model, the ID of the existing model must be provided
\item Whether to choose quick or accurate training
\end{itemize}

\subsubsection*{Key Points}
\begin{itemize}
\item If comprehensive evaluation or specialized evaluation is planned, it is recommended to choose accurate training. For model iteration, quick training can be chosen to improve efficiency.
\item In principle, training should continue from the previous model, but if it is observed that the MAE metric deteriorates rapidly, the cause may be a local minimum, in which case starting a new model training task from scratch can be considered.
\end{itemize}

\subsection*{\texttt{evaluation}}

\subsubsection*{What is the \texttt{evaluation} task}
The \texttt{evaluation} task involves evaluating the current machine-learned interatomic potential (MLIP) model to determine its stability and accuracy in molecular dynamics simulations. The \texttt{evaluation} task compares the differences between the current model and a reference method (such as PFP) in basic physical properties (density, RDF, MSD, etc.). After the \texttt{evaluation} task is completed, an evaluation report is usually generated for the decision maker to use in deciding the next action.

\subsubsection*{When to choose \texttt{evaluation}}
\begin{itemize}
\item In the most recent few rounds of \texttt{sample}, the MLIP MD has been stable (no divergence, no crash), and the training MAE has converged to a good level; otherwise, continue the \texttt{sample}/\texttt{train} cycle.
\end{itemize}

\subsubsection*{What information the task description should include}
\begin{itemize}
\item The type of structures that need to be evaluated
\item The ID of the model to use
\end{itemize}

\subsection*{\texttt{prune}}

\subsubsection*{What is the \texttt{prune} task}
The \texttt{prune} task involves deleting outliers data from the existing training dataset, removing one or more datasets that significantly reduce model reliablity and deleting thee unstable MLIPs.

\subsubsection*{When to choose \texttt{prune}}
\begin{itemize}
\item{MLIP performance in evaluation has deteriorated significantly}
\item{MLIP stability in the sampling task has deteriorated significantly.}
\end{itemize}

\subsubsection*{What information the task description should include}
\begin{itemize}
\item{Latest and second-latest model IDs} 
\item{Recently generated dataset file}
\end{itemize}

\subsection*{\texttt{End}}

\subsubsection*{What is the \texttt{end} task}
The \texttt{end} task involves terminating the entire active learning task and providing final recommendations and a summary. By ending the task, the decision maker can provide the user with recommendations regarding the performance of the MLIP model, its scope of applicability, and directions for future improvement.  
There are two exit paths for the task:
\begin{itemize}
\item Successful exit: when the MLIP model has reached the expected performance and stability, a successful exit can be chosen, and final recommendations can be provided.
\item Unsuccessful exit: when the MLIP model cannot reach the expected performance, or when the task encounters insurmountable difficulties, an unsuccessful exit can be chosen, and the reasons should be explained to the user.
\end{itemize}
After the \texttt{end} task is completed, a final report is usually generated for the user to refer to.

\subsubsection*{When to choose \texttt{end}}
\begin{itemize}
\item Successful exit. When the MLIP performs stably in the \texttt{sample} task, the number of samples selected in the \texttt{selection} task tends to stabilize, and reasonable performance metrics are achieved at all levels of \texttt{evaluation} tasks.
\item Unsuccessful exit. When the MLIP model has persistent problems, for example, the MLIP frequently fails in the \texttt{sample} task or cannot achieve even the most basic stability in the \texttt{evaluation} task. After multiple further attempts at model training and sample selection, the model performance still cannot be improved.
\end{itemize}

\subsubsection*{What information the task description should include}
\begin{itemize}
\item The type of exit (successful exit or unsuccessful exit).
\end{itemize}

\clearpage

\section{Package architecture}
\label{app:package_arch}

This appendix summarizes the package architecture of Lang2MLIP.

\newcommand{\pyfile}[1]{{\color{pyblue}\texttt{#1}}}
\newcommand{\mdfile}[1]{{\color{mdred}\texttt{#1}}}
\newcommand{\jsonfile}[1]{{\color{jsongreen}\texttt{#1}}}
\newcommand{\pydir}[1]{{\texttt{#1}}}

\dirtree{%
.1 \texttt{Lang2MLIP}.
.2 \texttt{active\_learning\_agents/}\DTcomment{Active learning multi-agent orchestration}.
.3 \pyfile{orchestrator.py}\DTcomment{Main orchestrator: decision routing, async execution}.
.3 \mdfile{global\_prompt.md}\DTcomment{Shared context for all agents}.
.3 \mdfile{decision\_maker\_prompt.md}\DTcomment{Decision Maker: selects next task}.
.3 \mdfile{pfp\_sample\_prompt.md}\DTcomment{PFP-based initial sampling}.
.3 \mdfile{sample\_prompt.md}\DTcomment{MD sampling task}.
.3 \mdfile{selection\_prompt.md}\DTcomment{Frame selection task}.
.3 \mdfile{train\_prompt.md}\DTcomment{Model training task}.
.3 \mdfile{evaluation\_prompt.md}\DTcomment{Model evaluation task}.
.3 \mdfile{eval\_reference\_prompt.md}\DTcomment{Reference MD evaluation task}.
.3 \mdfile{prune\_prompt.md}\DTcomment{Dataset pruning task}.
.3 \mdfile{end\_agent.md}\DTcomment{Termination task}.
.3 \texttt{template/}\DTcomment{Parameterized execution scripts}.
.4 \pyfile{run\_sample.py}\DTcomment{NVT-NPT MD with real-time anomaly detection}.
.4 \pyfile{run\_selection.py}\DTcomment{Frame selection: all / error-based}.
.4 \pyfile{run\_train.py}\DTcomment{Training job submission}.
.4 \pyfile{run\_evaluation.py}\DTcomment{Lattice/EOS/elastic/phonon/defect/MD evaluation}.
.4 \pyfile{run\_eval\_reference\_md.py}\DTcomment{Parallel reference MD evaluation}.
.4 \jsonfile{pfp\_dataset\_generation\_control.json}\DTcomment{PFP sampling config}.
.2 \texttt{prepare/}\DTcomment{Preparation pipeline}.
.3 \pyfile{cli.py}\DTcomment{CLI entry: four-stage pipeline}.
.3 \pyfile{conversation\_agent.py}\DTcomment{Interactive task specification collection}.
.3 \pyfile{structure\_plan\_agent.py}\DTcomment{Structure generation task planning}.
.3 \pyfile{structure\_gen\_agent.py}\DTcomment{Routing to specialized structure agents}.
.3 \pyfile{reference\_md\_agent.py}\DTcomment{Reference MD script generation}.
.3 \mdfile{global\_prompt.md}\DTcomment{Shared context for prepare agents}.
.3 \mdfile{conversation\_prompt.md}\DTcomment{Conversational task specification prompt}.
.3 \mdfile{structure\_plan\_prompt.md}\DTcomment{Structure planning prompt}.
.3 \mdfile{structure\_gen\_prompt.md}\DTcomment{Structure generation routing prompt}.
.3 \mdfile{reference-md-prompt.md}\DTcomment{Reference MD generation prompt}.
.3 \mdfile{solid-agent.md}\DTcomment{Bulk crystal structure agent}.
.3 \mdfile{molecule-agent.md}\DTcomment{Molecular / liquid structure agent}.
.3 \mdfile{cluster-agent.md}\DTcomment{Nanoparticle / cluster agent}.
.3 \mdfile{amorphous-agent.md}\DTcomment{Amorphous structure agent}.
.3 \mdfile{polymer-agent.md}\DTcomment{Polymer structure agent}.
.3 \mdfile{solid-surface-agent.md}\DTcomment{Surface slab agent}.
.3 \mdfile{solid-solid-agent.md}\DTcomment{Solid--solid interface agent}.
.3 \mdfile{solid-molecule-agent.md}\DTcomment{Solid--molecule interface agent}.
.3 \mdfile{liquid-liquid-agent.md}\DTcomment{Liquid--liquid interface agent}.
.3 \mdfile{other-agent.md}\DTcomment{Fallback / custom structure agent}.
.2 \texttt{utils/}\DTcomment{Shared utility library}.
.3 \pyfile{calculator\_base.py}\DTcomment{PFP / LightPFP calculator factory}.
.3 \pyfile{check\_convergence.py}\DTcomment{Convergence detection (std / slope / range)}.
.3 \pyfile{check\_dataset.py}\DTcomment{Dataset validation against reference values}.
.3 \pyfile{md\_utils.py}\DTcomment{NVT equilibration + NPT production MD}.
.3 \pyfile{model\_utils.py}\DTcomment{gRPC query for completed training jobs}.
.3 \pyfile{selection\_utils.py}\DTcomment{Per-frame error, parallel selection, H5 output}.
.3 \pyfile{structure\_check.py}\DTcomment{Nearest-neighbor distance with PBC}.
.3 \pyfile{train\_utils.py}\DTcomment{Training time estimation (linear model)}.
.3 \pyfile{trajectory\_tools.py}\DTcomment{Trajectory readability / frame-count tools}.
.3 \pyfile{eval\_plot\_utils.py}\DTcomment{Evaluation result plotting utilities}.
.3 \texttt{report\_utils/}\DTcomment{JSONL report system}.
.4 \pyfile{\_base.py}\DTcomment{Low-level JSONL read / write}.
.4 \pyfile{dataset\_report.py}\DTcomment{Dataset records (structure count, energy range)}.
.4 \pyfile{train\_report.py}\DTcomment{Training records (model ID, MAE, outliers)}.
.4 \pyfile{trajectory\_report.py}\DTcomment{Trajectory records (snapshots, MD conditions)}.
.3 \texttt{structures/}\DTcomment{Structure generation library}.
.4 \pyfile{utils.py}\DTcomment{Supercell / SMILES / density / liquid layer utils}.
.4 \pyfile{solid.py}\DTcomment{Materials Project query, element substitution}.
.4 \pyfile{cluster.py}\DTcomment{Cubic / spherical cluster cutting}.
.4 \pyfile{solid\_interface.py}\DTcomment{Crystal surface slab generation}.
.4 \pyfile{solid\_liquid\_interface.py}\DTcomment{Solid--liquid interface construction}.
.4 \pyfile{solid\_solid\_interface.py}\DTcomment{Solid--solid interface construction}.
.4 \pyfile{liquid\_liquid\_interface.py}\DTcomment{Liquid--liquid interface construction}.
.4 \pyfile{soak.py}\DTcomment{Molecule insertion into host structure}.
.4 \pyfile{rnemd\_utils.py}\DTcomment{RNEMD thermal conductivity / viscosity}.
}

\clearpage

\section{Active-Learning Workflow Log}
\label{app:workflow_log}

This appendix presents the interaction log between the decision-making agent and the action agents, illustrating how information is passed among agents throughout the workflow. It provides a clear step-by-step record of how the multi-agent system decomposes and solves the complex MLIP development task through sequential decisions, execution, and feedback.

\begin{itemize}
    \item \textbf{Step 1. DecisionMakerAgent:} Selected task \texttt{eval\_reference}. The task aims to use PFP to run MD simulations on validation structures, collect basic physical information, and generate reference trajectories for subsequent evaluation. This establishes the baseline for the active-learning workflow.
    
    \item \textbf{Step 1. EvalReferenceAgent:} Completed evaluation on 10 validation structures, all successful. Each structure was simulated with NPT MD at 300 K, generating 500 snapshots. The complete report is available at \texttt{eval\_reference/report.txt}.
    
    Key physical quantity ranges (density, energy per atom, minimum interatomic distance) are summarized below:
    \begin{itemize}
        \item \texttt{graphite\_xlarge\_300K}: $\rho=[2.23, 2.23]$ g/cm$^3$, $E/\mathrm{at}=[-7.79, -7.78]$ eV, $d_{\min}=1.32$ \AA
        \item \texttt{Li2CO3\_amorphous\_large\_300K}: $\rho=[1.67, 1.77]$ g/cm$^3$, $E/\mathrm{at}=[-5.12, -5.11]$ eV, $d_{\min}=1.12$ \AA
        \item \texttt{LEDC\_amorphous\_medium\_low\_density\_300K}: $\rho=[1.15, 1.39]$ g/cm$^3$, $E/\mathrm{at}=[-4.90, -4.89]$ eV, $d_{\min}=0.95$ \AA
        \item \texttt{EC\_DMC\_liquid\_medium\_seed2\_300K}: $\rho=[0.56, 0.88]$ g/cm$^3$, $E/\mathrm{at}=[-4.54, -4.53]$ eV, $d_{\min}=0.99$ \AA
        \item \texttt{LiPF6\_EC\_DMC\_electrolyte\_large\_300K}: $\rho=[0.60, 1.03]$ g/cm$^3$, $E/\mathrm{at}=[-4.52, -4.51]$ eV, $d_{\min}=1.00$ \AA
        \item \texttt{graphite\_1010\_slab\_07\_300K}: $\rho=[0.87, 0.88]$ g/cm$^3$, $E/\mathrm{at}=[-7.68, -7.67]$ eV, $d_{\min}=1.28$ \AA
        \item \texttt{graphite\_1010\_Li2CO3\_interface\_06\_300K}: $\rho=[1.82, 1.86]$ g/cm$^3$, $E/\mathrm{at}=[-6.81, -6.80]$ eV, $d_{\min}=1.14$ \AA
        \item \texttt{li2co3\_ledc\_interface\_lowdens\_300K}: $\rho=[1.39, 1.51]$ g/cm$^3$, $E/\mathrm{at}=[-5.05, -5.04]$ eV, $d_{\min}=1.02$ \AA
        \item \texttt{ledc\_electrolyte\_interface\_07\_300K}: $\rho=[0.92, 1.11]$ g/cm$^3$, $E/\mathrm{at}=[-4.72, -4.71]$ eV, $d_{\min}=1.01$ \AA
        \item \texttt{multilayer\_sei\_thick\_sei\_300K}: $\rho=[1.48, 1.53]$ g/cm$^3$, $E/\mathrm{at}=[-5.19, -5.18]$ eV, $d_{\min}=0.92$ \AA
    \end{itemize}
    
    \item \textbf{Step 2. DecisionMakerAgent:} Selected task \texttt{pfp\_sample}. The goal is to use PFP to comprehensively sample all non-validation initial structures and generate the initial training dataset, prioritizing basic components such as graphite, Li$_2$CO$_3$, LEDC, and electrolyte.
    
    \item \textbf{Step 2. PFPSampleAgent:} Completed initial PFP sampling and generated \texttt{selection/init.h5}. Sampling methods included MD (NPT at 300/600/1000 K, NVT at 1500/2000 K), compression ($0.92$--$1.08$), and \texttt{rattle\_and\_relax} (150 samples). A total of 20 non-validation initial structures were covered, producing 3312 snapshots and 506784 atoms.
    
    \item \textbf{Step 3. DecisionMakerAgent:} Selected task \texttt{train}. Since 3312 high-quality training snapshots had been generated, the next step was to train the first MLIP model from scratch.
    
    \item \textbf{Step 3. TrainAgent:} Completed Stage 1 first MLIP training from scratch. Model \texttt{04120005} (ID: \texttt{wma3pblet88e3d67}) was trained for 1237 epochs. Final errors were 0.011 eV/atom for energy and 0.33 eV/\AA{} for force. A total of 1501 anomalous data points (45\%) were detected, suggesting that targeted sample selection would be beneficial in the next iteration.
    
    \item \textbf{Step 4. DecisionMakerAgent:} Selected task \texttt{sample}. The newly trained MLIP model was used to test MD stability on Stage 1 basic-component structures over 300--1000 K.
    
    \item \textbf{Step 4. SampleAgent:} All 48 MD tasks completed successfully with no early stops or anomalies. The MLIP model \texttt{wma3pblet88e3d67} showed excellent stability in the 300--1000 K range, with no energy divergence, structural collapse, density anomalies, or temperature runaway. A total of 4848 snapshots were collected from 48 trajectories. The sampled systems covered graphite (12 tasks, 300--1000 K), Li$_2$CO$_3$ (12 tasks, 300--1000 K), LEDC (12 tasks, 300--900 K), and LiPF$_6$/EC/DMC electrolyte (12 tasks, 300--800 K).
    
    \item \textbf{Step 5. DecisionMakerAgent:} Selected task \texttt{selection}. Since the model showed excellent stability, 5--10\% of the sampled configurations were recommended for reference calculations.
    
    \item \textbf{Step 5. SelectionAgent:} Completed sample selection (task ID: \texttt{04120404}). From 4848 snapshots, the top 7.5\% high-error samples were selected using the \texttt{max\_force} criterion, resulting in \texttt{selection/dataset\_04120404.h5} with 364 structures (40730 atoms). The energy range was $-5.01$ to $-4.44$ eV/atom (mean $-4.75$), and the maximum force was 16.94 eV/\AA. Overall model performance was strong, with average energy MAE of 0.032 eV/atom and force MAE of 0.21 eV/\AA.
    
    \item \textbf{Step 6. DecisionMakerAgent:} Selected task \texttt{train}. The existing model \texttt{wma3pblet88e3d67} was further refined with the newly selected 364 high-error samples.
    
    \item \textbf{Step 6. TrainAgent:} Completed MLIP fine-tuning. Model \texttt{04120407} (ID: \texttt{ubjddsyv65chhyfd}) was trained from \texttt{wma3pblet88e3d67} using 3676 structures (3312 initial + 364 new samples) for 2000 epochs. Final errors were 0.014 eV/atom for energy and 0.322 eV/\AA{} for force. Six anomalous data points were found in \texttt{init.h5}, while no anomalies appeared in the new dataset.

    \item \textbf{Step 7. DecisionMakerAgent:} Selected task \texttt{sample}. Model \texttt{ubjddsyv65chhyfd} was used to begin Stage 2 interface sampling, focusing on graphite/Li$_2$CO$_3$ and Li$_2$CO$_3$/LEDC interfaces in the 300--800 K range.
    
    \item \textbf{Step 7. SampleAgent:} Sampling was partially completed, with 58 out of 72 tasks successful and 14 early stops. All 42 Li$_2$CO$_3$/LEDC interface tasks succeeded, whereas 14 out of 30 graphite/Li$_2$CO$_3$ tasks stopped early due to structural collapse at high temperatures; 9 of these occurred at temperatures $\ge 700$ K. A total of 13149 snapshots were obtained.
    
    \item \textbf{Step 8. DecisionMakerAgent:} Selected task \texttt{selection}. Based on the sampling results, high-error samples and early-stop trajectories from graphite/Li$_2$CO$_3$ were prioritized.
    
    \item \textbf{Step 8. SelectionAgent:} Completed Stage 2 interface sample selection, producing two datasets:
    \begin{enumerate}
        \item graphite/Li$_2$CO$_3$ (17.5\% sampling): 824 structures, 273370 atoms, energy range $-6.50$ to $-5.14$ eV/atom, maximum force 379.86 eV/\AA
        \item Li$_2$CO$_3$/LEDC (7.5\% sampling): 633 structures, 162514 atoms, energy range $-5.00$ to $-4.81$ eV/atom, maximum force 15.34 eV/\AA
    \end{enumerate}
    In total, 1457 high-error samples were selected.
    
    \item \textbf{Step 9. DecisionMakerAgent:} Selected task \texttt{train}. The model was fine-tuned using the 1457 interface high-error samples.
    
    \item \textbf{Step 9. TrainAgent:} Completed fine-tuning of model \texttt{04122324} (ID: \texttt{ryhw6lnfvitqypsn}) based on \texttt{ubjddsyv65chhyfd}. After 2000 epochs, the energy MAE decreased by 21.5\% to 0.0107 eV/atom, and the force MAE decreased by 3.8\% to 0.310 eV/\AA. Eighteen anomalous data points were detected, 10 of which came from graphite/Li$_2$CO$_3$.
    
    \item \textbf{Step 10. DecisionMakerAgent:} Selected task \texttt{sample}. The updated model \texttt{ryhw6lnfvitqypsn} was used to re-sample Stage 2 interfaces, especially to test whether high-temperature stability at the graphite/Li$_2$CO$_3$ interface had improved.
    
    \item \textbf{Step 10. SampleAgent:} All 62 MD tasks completed successfully with no early stops or anomalies. The new model showed stable graphite/Li$_2$CO$_3$ behavior at 700--1000 K, in contrast to the previous 14/30 early stops. Li$_2$CO$_3$/LEDC remained stable over 300--1000 K. A total of 18662 snapshots were collected.
    
    \item \textbf{Step 11. DecisionMakerAgent:} Selected task \texttt{selection}. Since the model exhibited excellent stability, 5--10\% of samples were recommended for reference calculations.
    
    \item \textbf{Step 11. SelectionAgent:} From 18662 snapshots, 1400 high-error samples (7.5\%) were selected, covering both graphite/Li$_2$CO$_3$ and Li$_2$CO$_3$/LEDC interfaces over 300--1000 K. The maximum force error reached 39 eV/\AA{} for graphite/Li$_2$CO$_3$ and 25 eV/\AA{} for Li$_2$CO$_3$/LEDC.
    
    \item \textbf{Step 12. DecisionMakerAgent:} Selected task \texttt{train}. The model was further refined with these 1400 Stage 2 interface samples.
    
    \item \textbf{Step 12. TrainAgent:} Completed fine-tuning of model \texttt{04131346} (ID: \texttt{o5iksgkqtw89x4pt}). Training used all five datasets, including the 1400 newly selected interface samples, over 1580 epochs. Fifteen anomalous data points were detected (6 from \texttt{init.h5}, 9 from the graphite/Li$_2$CO$_3$ dataset). Final training errors were 0.0097 eV/atom for energy and 0.312 eV/\AA{} for force.
    
    \item \textbf{Step 13. DecisionMakerAgent:} Selected task \texttt{evaluation}. The latest model \texttt{o5iksgkqtw89x4pt} was used to evaluate the graphite/Li$_2$CO$_3$ and Li$_2$CO$_3$/LEDC interfaces.
    
    \item \textbf{Step 13. EvaluationAgent:} Completed Stage 2 interface evaluation. For the graphite/Li$_2$CO$_3$ interface, the density deviation was 0.6\%, and both RDF and energy curves were highly consistent with PFP. For the Li$_2$CO$_3$/LEDC interface, the density deviation was 7.6\%, the RDF main-peak positions matched well, although the predicted volume was slightly smaller. No structural collapse was observed in either case.

    \item \textbf{Step 14. DecisionMakerAgent:} Selected task \texttt{sample}. The model \texttt{o5iksgkqtw89x4pt} was applied to the final Stage 2 interface type, namely the LEDC/electrolyte interface, in order to complete interface coverage.
    
    \item \textbf{Step 14. SampleAgent:} Sampling completed with 35 out of 36 MD tasks successful; one task stopped early at 600 K due to a sudden temperature increase to 1574 K. Overall success rate was 97\%. A total of 10823 snapshots were collected from six LEDC/electrolyte interface structures over 300--800 K.
    
    \item \textbf{Step 15. DecisionMakerAgent:} Selected task \texttt{selection}. Because the model performed well, 5--10\% of the sampled structures were recommended for reference calculations.
    
    \item \textbf{Step 15. SelectionAgent:} From 10823 snapshots, 812 high-error structures (7.5\%) were selected. The resulting dataset \texttt{selection/dataset\_04140121.h5} contained 448400 atoms. The energy range was $-4.68$ to $-4.52$ eV/atom, the average force MAE was 0.23 eV/\AA, and the maximum force error was 15.98 eV/\AA.
    
    \item \textbf{Step 16. DecisionMakerAgent:} Selected task \texttt{train}. The model was refined with these 812 LEDC/electrolyte samples.
    
    \item \textbf{Step 16. TrainAgent:} Completed fine-tuning of model \texttt{04140126} (ID: \texttt{ie4o21a3w3xxr0qo}) based on \texttt{o5iksgkqtw89x4pt}. Fourteen anomalous data points were found (4 from \texttt{init.h5}, 10 from graphite/Li$_2$CO$_3$). Final training errors were 0.0085 eV/atom for energy and 0.275 eV/\AA{} for force, corresponding to an 11.8\% improvement in force error over the base model.

    \item \textbf{Step 17. DecisionMakerAgent:} Selected task \texttt{sample}. The model \texttt{ie4o21a3w3xxr0qo} was used to sample complete multilayer SEI structures (graphite/Li$_2$CO$_3$/LEDC/electrolyte) over 300--600 K.
    
    \item \textbf{Step 17. SampleAgent:} All 8 MD tasks completed successfully with no early stops or anomalies. The model exhibited stable performance for the complete four-layer SEI system, with no energy divergence, structural collapse, or temperature runaway. A total of 2408 snapshots were obtained. The sampled systems included two multilayer SEI structures: a compact system with 914 atoms and a standard system with 1153 atoms.
    
    \item \textbf{Step 18. DecisionMakerAgent:} Selected task \texttt{selection}. Since the model performed stably, 5--10\% of the snapshots were recommended for reference calculations.
    
    \item \textbf{Step 18. SelectionAgent:} Completed sample selection for the four-layer SEI interface, generating \texttt{selection/dataset\_04140836.h5} with 181 structures (approximately 207k atoms). The energy range was $-5.253$ to $-5.087$ eV/atom. The maximum force was 10.647 eV/\AA. The energy MAE ranged from 0.0034 to 0.0101 eV/atom, while the force MAE ranged from 0.1940 to 0.2176 eV/\AA.
    
    \item \textbf{Step 19. DecisionMakerAgent:} Selected task \texttt{train}. The model was further refined using these 181 multilayer SEI samples.
    
    \item \textbf{Step 19. TrainAgent:} Completed fine-tuning of model \texttt{04140839} (ID: \texttt{vhdko7lyhyhtbqqd}) based on \texttt{ie4o21a3w3xxr0qo}. After adding the 181 multilayer samples and training for 1073 epochs, the force MAE decreased from 0.275 to 0.262 eV/\AA{} (a 4.7\% improvement), while the energy MAE remained at 0.0085 eV/atom. No anomalies were found in the new dataset.

    \item \textbf{Step 20. DecisionMakerAgent:} Selected task \texttt{evaluation}. The final model \texttt{vhdko7lyhyhtbqqd} was used for comprehensive evaluation on the complete multilayer SEI structure.
    
    \item \textbf{Step 20. EvaluationAgent:} Completed the comprehensive multilayer SEI evaluation. The MLIP model \texttt{vhdko7lyhyhtbqqd} was highly consistent with PFP, showing a density deviation of only 0.70\%. MD remained stable over 50000 steps with no structural collapse. RDF curves for all atom pairs (e.g., H--H, Li--Li, C--O, and F--F) matched closely, and the MSD indicated reasonable diffusion behavior. Energy, density, and volume fluctuations were normal. The minimum interatomic distance remained above 0.93 \AA, and the maximum force was below 5.9 eV/\AA.
    
    \item \textbf{Step 21. DecisionMakerAgent:} Selected task \texttt{end}. The workflow was successfully completed. The final MLIP model \texttt{vhdko7lyhyhtbqqd} passed all Stage 1 and Stage 2 training and evaluation steps, achieving strong consistency with PFP on multilayer SEI structures. The final model metrics were 0.0085 eV/atom for energy MAE and 0.262 eV/\AA{} for force MAE. The model was considered ready for Stage 3 Li$^+$ crossing metadynamics simulations. The recommended setup is to combine this model with Plumed in well-tempered metadynamics, using the Li$^+$ $z$-coordinate as the collective variable, Gaussian height 0.5--1.0 kJ/mol, width 0.02 nm, bias factor 15--20, deposition interval 200--500 fs, total simulation time 5--10 ns, and temperature 300 K.
    
\end{itemize}

\appendix


\end{document}